\documentclass[runningheads]{llncs}

\usepackage{accv}
\usepackage{accvabbrv}

\usepackage{graphicx}
\usepackage{booktabs}
\usepackage{tabularx}
\usepackage{multirow}
\usepackage{siunitx}
\usepackage[accsupp]{axessibility}

\usepackage{hyperref}
\usepackage{orcidlink}

\usepackage[nomain,acronym]{glossaries}
\glsdisablehyper
\newacronym{kl}{KL}{Kernel Linking}
\newacronym{gfa}{GFA}{Group Feature Assembling}
\newacronym{aku}{AKU}{Adaptive Kernel Update}
\newacronym{ki}{KI}{Kernel Interaction}
\newacronym{mhsa}{MHSA}{Multi-Head Self-Attention}
\newacronym{ffn}{FFN}{Feed-Forward Neural Network}
\newacronym{iou}{IoU}{intersection over union}
\newacronym{ssim}{SSIM}{Structured Similarity Index}
\newacronym{pq}{PQ}{Panoptic Quality}

\newcolumntype{Y}{>{\centering\arraybackslash}X}

\begin{document}

\title{MGNiceNet: Unified Monocular Geometric Scene Understanding}

\author{Markus Schön\orcidlink{0009-0006-6029-9341} \and
Michael Buchholz\orcidlink{0000-0001-5973-0794} \and
Klaus Dietmayer\orcidlink{0000-0002-1651-014X}%
}

\authorrunning{M. Schön~\etal}

\institute{Institute of Measurement, Control, and Microtechnology, Ulm University, Germany
\email{\{markus.schoen,michael.buchholz,klaus.dietmayer\}@uni-ulm.de}}

\maketitle

\begin{abstract}
  Monocular geometric scene understanding combines panoptic segmentation and self-supervised depth estimation, focusing on real-time application in autonomous vehicles.
  We introduce MGNiceNet, a unified approach that uses a linked kernel formulation for panoptic segmentation and self-supervised depth estimation.
  MGNiceNet is based on the state-of-the-art real-time panoptic segmentation method RT-K-Net and extends the architecture to cover both panoptic segmentation and self-supervised monocular depth estimation.
  To this end, we introduce a tightly coupled self-supervised depth estimation predictor that explicitly uses information from the panoptic path for depth prediction.
  Furthermore, we introduce a panoptic-guided motion masking method to improve depth estimation without relying on video panoptic segmentation annotations.
  We evaluate our method on two popular autonomous driving datasets, Cityscapes and KITTI.
  Our model shows state-of-the-art results compared to other real-time methods and closes the gap to computationally more demanding methods.
  Source code and trained models are available at \url{https://github.com/markusschoen/MGNiceNet}.
  
  \keywords{Monocular Geometric Scene Understanding \and Panoptic Segmentation \and Self-supervised Depth Estimation \and Multi-task Learning}
\end{abstract}
\section{Introduction}
\label{sec:intro}

Panoptic segmentation~\cite{kirillov2019panoptic1} was introduced to unify the tasks of semantic segmentation and instance segmentation by segmenting both amorph stuff regions and countable thing objects.
Stuff regions encompass semantic categories such as road or vegetation, while thing objects encompass instance categories such as pedestrian or vehicle.
This information is highly relevant for downstream tasks such as trajectory planning to navigate through the vehicle's environment safely.
The major drawback of panoptic segmentation for its application in autonomous driving systems is its limitation to the 2D image plane.
Depth information is essential to lift 2D information into our 3D world.
Therefore, it comes naturally to combine panoptic segmentation with monocular depth estimation~\cite{eigen2014depth}, which aims to recover the distance to the camera for each pixel in the image.
Both tasks share similarities,~\eg, depth discontinuities at object borders, which can be exploited.
Combined methods~\cite{gao2022panopticdepth, he2023towards, schoen2021mgnet, wang2020sdc, yuan2022polyphonicformer} often show improved results compared to single-task methods.
However, most of these methods share two main limitations for wide application in autonomous driving systems.
First, depth estimation is trained in a supervised way using either stereo disparities or projected lidar points as ground truth.
Supervised methods have limited generalization ability to different camera setups, and data acquisition for large-scale depth datasets is challenging.
It is time-consuming, and sensors need to be well-calibrated.
Second, most methods focus on high accuracy rather than a fast inference speed.
Methods need a reasonable inference speed to process incoming images quickly and keep the latency in the autonomous driving system low.
Otherwise, the system cannot react to sudden changes in the vehicle's environment,~\eg, pedestrians crossing the street.

In our previous work~\cite{schoen2021mgnet}, we introduced monocular geometric scene understanding,~\ie, a combination of panoptic segmentation and self-supervised monocular depth estimation~\cite{zhou2017unsupervised}, to tackle both of these limitations.
The self-supervised approach to depth estimation alleviates the problem of acquiring per-pixel ground truth depth using an image-to-image synthesis formulation, which requires only captured video sequences during training.
Thus, self-supervised methods can easily be adapted to new cameras by retraining the model with videos captured from the new camera.
Furthermore, we introduced MGNet, a network architecture which focuses on low latency for real-time applications.
Despite showing reasonable performance, MGNet has one main limitation: 
The task combination is only handled implicitly using a shared encoder and separate task-specific decoders.
The implicit formulation limits the model from exploiting the inherent task relations between panoptic segmentation and self-supervised depth estimation, ultimately limiting model performance.

Motivated by this limitation, we propose MGNiceNet, a unified monocular geometric scene understanding method that explicitly exploits the relations between the two tasks.
MGNiceNet is based on our state-of-the-art real-time panoptic segmentation method RT-K-Net~\cite{schoen2023rt}, which we extend to additionally perform the task of self-supervised depth estimation.
Specifically, we introduce a set of self-supervised depth kernels and employ depth kernel updates similar to PolyphonicFormer~\cite{yuan2022polyphonicformer} to produce instance-wise depth predictions.
Inspired by the recent success of classification-regression approaches to depth estimation~\cite{bhat2021adabins, wang2023sqldepth}, MGNiceNet also predicts depth using a discrete bin-wise depth representation.
However, compared to~\cite{bhat2021adabins, wang2023sqldepth}, we propose to use a bin-wise representation at the panoptic mask level, allowing for a better alignment between the panoptic segmentation and depth prediction.
Assumptions for self-supervised depth learning are a moving camera in a static environment and no occlusions between frames~\cite{zhou2017unsupervised}.
Thus, dynamic objects break this assumption, resulting in high photometric errors and incorrect depth predictions.
Several methods successfully incorporated motion masking techniques into self-supervised training~\cite{klingner2020self, petrovai2023monodvps}.
However, these methods either require a static frame selection~\cite{klingner2020self} or video panoptic annotations~\cite{petrovai2023monodvps}.
Thus, we propose a novel panoptic-guided motion masking method that uses panoptic predictions instead of video panoptic annotations.
In short, our main contributions can be summarized as follows:
\begin{itemize}
    \item 
    We introduce MGNiceNet, a unified method for monocular geometric scene understanding that extends our state-of-the-art real-time panoptic segmentation method RT-K-Net~\cite{schoen2023rt} with self-supervised depth estimation using explicit kernel linking between panoptic and depth kernels.
    \item 
    We propose a bin-wise depth representation at the panoptic mask level using a depth predictor module with low computational overhead.
    \item 
    We introduce a novel panoptic-guided motion masking technique based on warped panoptic segmentation masks with tracked instances, which do not need static frame selection or video panoptic annotations.
    \item 
    We evaluate our method on two standard autonomous driving benchmarks, Cityscapes~\cite{cordts2016cityscapes} and KITTI~\cite{geiger2013vision}, and show the effectiveness of our contributions in extensive experiments.
\end{itemize}
\section{Related Work}
\label{sec:rel_work}

\subsection{Panoptic Segmentation}
Panoptic segmentation~\cite{kirillov2019panoptic1} was introduced to unify semantic segmentation and instance segmentation.
Early works~\cite{kirillov2019panoptic2, porzi2019seamless, mohan2021efficientps, cheng2020panoptic, chen2020scaling} often treated panoptic segmentation as a multi-task problem and used a shared encoder with separate decoders for semantic segmentation and instance segmentation.
Recently, unified panoptic segmentation approaches~\cite{wang2021max, cheng2021per, zhang2021k, yu2022cmt, yu2022k, zhang2023unidaformer, jain2023oneformer} emerged using a mask classification approach to tackle both semantic segmentation and instance segmentation in a unified way.
K-Net~\cite{zhang2021k} was one of the first unified approaches.
It uses learnable kernels to produce mask proposals for both stuff and thing classes, which can be used to predict panoptic segmentation maps directly.
Current state-of-the-art methods such as OneFormer~\cite{jain2023oneformer} use transformer queries instead of learnable kernels, a multi-scale transformer decoder, and a task-conditioned text encoder to produce high-quality panoptic segmentation maps.
While such methods achieve high accuracy, they often rely on heavy architectures, making them unsuitable for real-time applications.

Only a few panoptic segmentation methods focus on real-time application~\cite{petrovai2020real, hou2020real, chen2020scaling, schoen2023rt}.
While most methods~\cite{petrovai2020real, hou2020real, chen2020scaling} rely on specialized formulations for semantic segmentation and instance segmentation, our previously introduced unified approach RT-K-Net~\cite{schoen2023rt} achieves state-of-the-art performance in real-time panoptic segmentation.
Thus, MGNiceNet is based on RT-K-Net, which we extend to the task of self-supervised depth estimation.

\subsection{Self-supervised Monocular Depth Estimation}
SfmLearner~\cite{zhou2017unsupervised} was the first method to use monocular video sequences to train a depth estimation model in a self-supervised manner.
Godard~\etal~\cite{godard2019digging} propose to upsample the multi-scale depth maps before loss calculation and use the minimum photometric error to tackle occlusions.
Furthermore, they propose an auto-masking strategy to avoid holes of infinite depth in low-texture regions and regions with dynamic objects.
Their contributions advanced the field considerably and were integrated by many recent methods~\cite{xue2020hierarchical, guizilini20203d, watson2021temporal, zhao2022monovit, wang2023sqldepth}, and we also build our framework on their foundation.
Wang~\etal~\cite{wang2023sqldepth} recently introduced a Self Query Layer to learn geometric object features from pixel features.
The features are then decoded using a bin-wise discrete depth representation at the geometric object level.
In contrast to~\cite{wang2023sqldepth}, we propose to predict depth bins at the panoptic mask level, which better aligns with the panoptic prediction without the need for the computationally heavy Self Query Layer.

\subsection{Combined Approaches}
Some works focus on incorporating semantic~\cite{guizilini2020semantically, klingner2020self} or panoptic~\cite{saeedan2021boosting} annotations to boost the performance of self-supervised depth maps.
For example, Klingner~\etal~\cite{klingner2020self} propose a multi-task model that performs semantic segmentation and self-supervised depth estimation.
They use a motion masking technique based on the predicted semantic segmentation map to exclude dynamic objects from the photometric loss calculation.
While they show the benefits of their masking technique, their approach is limited since object information is missing in the semantic segmentation prediction.
Other works combine depth estimation with image~\cite{xu2018pad, zhang2018joint, wang2020sdc, gao2022panopticdepth, he2023towards} or video panoptic segmentation~\cite{qiao2021vip, petrovai2023monodvps, ji_yeon2023mindvps, yuan2022polyphonicformer}.
Yuan~\etal~\cite{yuan2022polyphonicformer} introduce PolyphonicFormer, which uses query linking to boost depth estimation performance.
Our method leverages kernel linking and combines it with a depth predictor for the self-supervised training objective.
Petrovai and Nedevschi~\cite{petrovai2023monodvps} propose a combination of video panoptic segmentation and self-supervised depth estimation.
They also propose a panoptic-guided motion masking technique, but their method relies on video panoptic annotations, which have a high annotation cost.
In contrast, our motion masking technique only requires video sequence images since we calculate motion masks based on panoptic predictions.
Our previous approach MGNet~\cite{schoen2021mgnet} combines panoptic segmentation with self-supervised depth estimation focusing on real-time application.
However, MGNet uses separate task decoders for semantic segmentation, instance segmentation, and self-supervised depth estimation.
For MGNiceNet, we propose to use a unified decoder approach using linked kernel update heads.
\section{Method}
\label{sec:method}
\begin{figure}[tb]
  \centering
  \includegraphics[width=\textwidth]{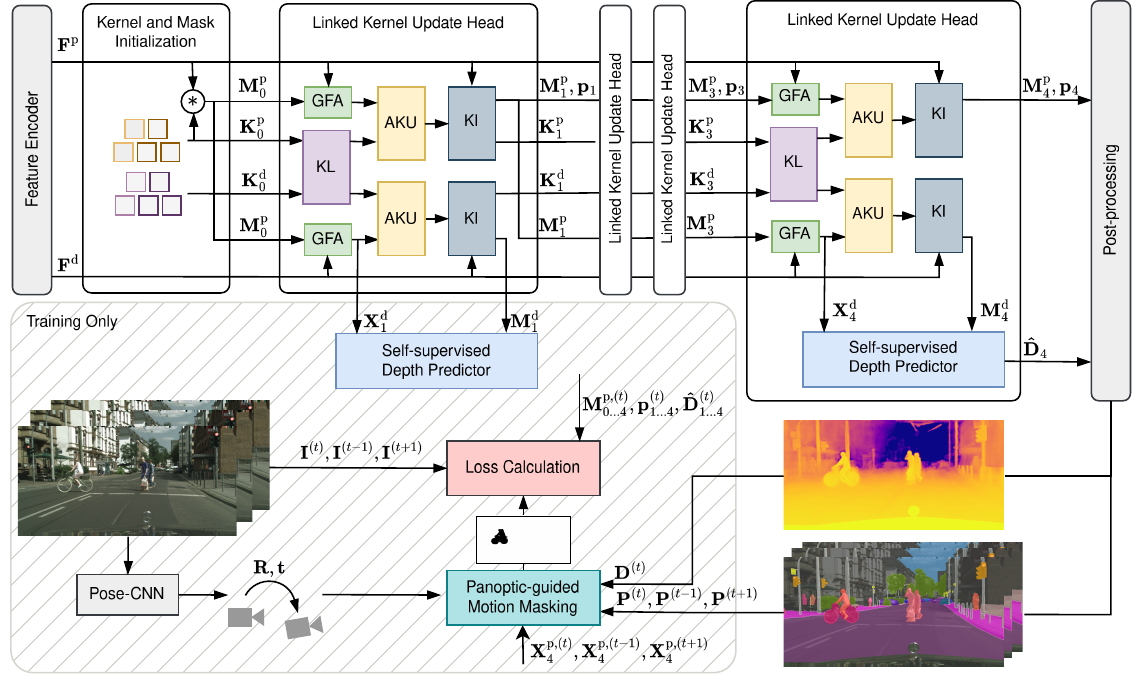}
  \caption{\textbf{Overview of our MGNiceNet architecture.} Images are fed into the Feature Encoder (\cref{sec:feat_enc}) to produce high-resolution feature maps $\mathbf{F}^{\mathrm{p}}$ and $\mathbf{F}^{\mathrm{d}}$. Next, the panoptic kernels $\mathbf{K}^{\mathrm{p}}_0$, self-supervised depth kernels $\mathbf{K}^{\mathrm{d}}_0$, and the panoptic masks $\mathbf{M}^{\mathrm{p}}_0$ are initialized (\cref{sec:mask_init}). Kernels and masks are updated iteratively in the four linked kernel update heads (\cref{sec:linked_kernels}). Each head contains \gls{kl}, \gls{gfa}, \gls{aku}, and \gls{ki} stages and a self-supervised depth predictor (\cref{sec:depth_pred}), which converts depth mask predictions $\mathbf{M}^{\mathrm{d}}_i$ and depth group features $\mathbf{X}^{\mathrm{d}}_i$ into an inverse depth prediction $\mathbf{\hat{D}}^{\mathrm{d}}_i$. A post-processing stage (\cref{sec:post_proc}) converts mask predictions $\mathbf{M}^{\mathrm{p}}_4$, class probability logits $\mathbf{p}_4$, and inverse depth predictions $\mathbf{\hat{D}}^{\mathrm{d}}_4$ into the final panoptic prediction $\textbf{P}$ and depth prediction $\mathbf{D}$. To improve optimization (\cref{sec:optimization}), we introduce a panoptic-guided motion masking method (\cref{sec:motion_mask}), which calculates a motion mask to mask out dynamic objects from the photometric loss.}
  \label{fig:overview}
\end{figure}

\subsection{Problem Formulation}
\label{sec:problem}
Given a batch of $B$ input color images $\mathbf{I}^{(t)}\in\mathbb{R}^{B\times 3 \times H \times W}$ at time step $t$ and spatial dimensions $H$ and $W$ the task of monocular geometric scene understanding is to infer a panoptic prediction $\mathbf{P}\in\mathbb{N}^{B\times H \times W}$ and a depth prediction $\mathbf{D}\in\mathbb{R}^{B\times H \times W}$.
The panoptic prediction assigns a unique class ID of a predefined number of classes $N_c$ to each pixel in the image and a unique instance ID to each pixel corresponding to thing classes.
The depth prediction assigns a metric depth value,~\ie, the metric distance to the camera, to each pixel in the image.
Panoptic segmentation can be formulated as a mask classification problem~\cite{cheng2021per}.
The idea is to predict a fixed number of $N$ mask predictions $\mathbf{M}^{p}\in\mathbb{R}^{B\times N \times H \times W}$ with corresponding class probability predictions $\mathbf{p}\in\mathbb{R}^{B\times N \times N_c}$.
For stuff classes, a single mask per class is predicted, while each thing class can have multiple mask predictions depending on the number of objects
present in the input image.
During training, probability-mask pairs are assigned to ground truth using bipartite matching~\cite{cheng2021per}.
Depth estimation can be formulated in a self-supervised way as an image-to-image synthesis problem using monocular video sequences~\cite{zhou2017unsupervised}.
During training, based on the input reference frame $\mathbf{I}^{(t)}$ at time step $t$, a depth prediction $\mathbf{D}^{(t)}$ is used in conjunction with relative camera pose predictions to project adjacent video sequence frames,~\ie, context frames $\mathbf{I}_c\in\{\mathbf{I}^{(t-1)}, \mathbf{I}^{(t+1)}\}$ into the reference frame.
Optimization uses a photometric image reconstruction loss to maximize the similarity between the warped context frames and the reference frame.

\subsection{Method Overview}
\label{sec:method_overview}
An overview of our method is given in \cref{fig:overview}.
MGNiceNet tackles monocular geometric scene understanding in a unified framework.
The network architecture is built upon RT-K-Net~\cite{schoen2023rt}, which is extended to self-supervised depth estimation. 
The following sections detail the proposed extension of the network architecture, the optimization process, and the proposed panoptic-guided motion masking.

\subsection{Feature Encoder}
\label{sec:feat_enc}
Following RT-K-Net, RTFormer~\cite{wang2022rtformer} is used as a feature encoder.
RTFormer uses the input image batch to generate a single high-resolution feature map $\mathbf{F}^{\mathrm{p}}\in\mathbb{R}^{B\times C \times H' \times W'}$ with feature channel size $C$ and $H'=\frac{H}{8}, W'=\frac{W}{8}$.
For MGNiceNet, a second feature map $\mathbf{F}^{\mathrm{d}}\in\mathbb{R}^{B\times C \times H' \times W'}$ with the same dimensions is generated by adding a parallel convolution stage to the last encoder layer.

\subsection{Kernel and Mask Initialization}
\label{sec:mask_init}
RT-K-Net uses randomly initialized panoptic kernels $\mathbf{K}^{\mathrm{p}}_0\in\mathbb{R}^{B\times N \times C}$ to produce initial panoptic mask predictions by performing convolution $\mathbf{M}^{\mathrm{p}}_0 = \mathbf{K}^{\mathrm{p}}_0 \ast \mathbf{F}^{\mathrm{p}}$ using the initialized panoptic kernels and the panoptic feature map from the feature encoder.
For MGNiceNet, self-supervised depth kernels $\mathbf{K}^{\mathrm{d}}_0\in\mathbb{R}^{B\times N \times C}$ are initialized similar to panoptic kernels.
However, no depth mask predictions $\mathbf{M}^{\mathrm{d}}_0$ are initialized since panoptic masks are used to refine the self-supervised depth kernels in the linked kernel update heads instead.

\subsection{Linked Kernel Update Heads}
\label{sec:linked_kernels}
The initial kernels and mask predictions are not discriminative enough to perform panoptic segmentation~\cite{zhang2021k}.
Hence, RT-K-Net performs four iterative kernel updates to increase the representation ability of each kernel.
The same holds for self-supervised depth kernels.
Therefore, the kernel update heads of RT-K-Net are extended to update the self-supervised depth kernels.
Each linked kernel update head $i$ performs the following stages to update kernels and mask predictions from the last linked kernel update head $i-1$:

\textbf{\acrfull{kl}} is performed similarly to query linking introduced in~\cite{yuan2022polyphonicformer} to update self-supervised depth kernels by reusing the information present in panoptic kernels.
The linking is unidirectional,~\ie, information is only passed from panoptic to self-supervised depth kernels using $\mathbf{\tilde{K}}^{\mathrm{d}}_i = \mathbf{K}^{\mathrm{d}}_{i-1} + \mathbf{K}^{\mathrm{p}}_{i-1}$.

\textbf{\acrfull{gfa}} is used to reduce noise and speed up the learning process~\cite{zhang2021k}.
The panoptic mask predictions $\mathbf{M}^{\mathrm{p}}_{i-1}$ are first converted to binary mask predictions $\mathbf{S}^{\mathrm{p}}_{i-1} = \sigma(\mathbf{M}^{\mathrm{p}}_{i-1}) > 0.5$.
Then, panoptic group features $\mathbf{X}^{\mathrm{p}}_i\in\mathbb{R}^{B\times N \times C}$ are formed using element-wise multiplication of the panoptic feature map $\mathbf{F}^{\mathrm{p}}$ and the normalized binary mask predictions
\begin{equation}
	\mathbf{X}^{\mathrm{p}}_i = \sum_{u}^{H'}\sum_{v}^{W'} \frac{\mathbf{S}^{\mathrm{p}}_{i-1}(u, v)}{\sum_{H'W'}\mathbf{S}^{\mathrm{p}}_{i-1}(u, v)} \mathbf{F}^{\mathrm{p}}(u, v).
\label{eq:gfa}
\end{equation}
We use RT-K-Net's normalization strategy to enable mixed-precision training and inference, which has been proven to reduce inference time~\cite{schoen2023rt}.
Depth group features $\mathbf{X}^{\mathrm{d}}_i\in\mathbb{R}^{B\times N \times C}$ are generated similarly to~\cref{eq:gfa} using the depth feature map $\mathbf{F}^{\mathrm{d}}$ and the normalized binarized mask predictions $\mathbf{S}^{\mathrm{p}}_{i-1}$.
This creates another link from the panoptic path to the depth path.

\textbf{\acrfull{aku}} is used to re-weight the influence of panoptic kernels and panoptic group features~\cite{zhang2021k} using learned gates $G_{\mathbf{F}}, G_{\mathbf{K}}$ and linear transformations $\psi_1, \psi_2$.
\gls{aku} is computationally inexpensive since it is independent of the image resolution.
Thus, similar to~\cite{yuan2022polyphonicformer}, we use two parallel \gls{aku} modules to update the panoptic and self-supervised depth kernels
\begin{align}
    \mathbf{\tilde{K}}^{\mathrm{p}}_i &= {G}^{\mathrm{p}}_{\mathbf{F}} \otimes \psi^{\mathrm{p}}_1(\mathbf{X}^{\mathrm{p}}_i)  + {G}^{\mathrm{p}}_{\mathbf{K}} \otimes \psi^{\mathrm{p}}_2(\mathbf{K}^{\mathrm{p}}_{i-1}) \\
    \mathbf{\tilde{K}}^{\mathrm{d}}_i &= {G}^{\mathrm{d}}_{\mathbf{F}} \otimes \psi^{\mathrm{d}}_1(\mathbf{X}^{\mathrm{d}}_i)  + {G}^{\mathrm{d}}_{\mathbf{K}} \otimes \psi^{\mathrm{d}}_2(\mathbf{\tilde{K}}^{\mathrm{d}}_i).
\end{align}

\textbf{\acrfull{ki}} is performed to model the relationship between kernels, \ie, between different pixel groups.
Similar to~\gls{aku}, this operation is inexpensive and thus extended by adding a second parallel \gls{ki} module to update self-supervised depth kernels
\begin{equation}
    \mathbf{K}^{\mathrm{p}}_i = \mathrm{FFN}^{\mathrm{p}}_{\mathbf{K}}(\mathrm{MHSA}^{\mathrm{p}}(\mathbf{\tilde{K}}^{\mathrm{p}}_i)),\quad \mathbf{K}^{\mathrm{d}}_i = \mathrm{FFN}^{\mathrm{d}}_{\mathbf{K}}(\mathrm{MHSA}^{\mathrm{d}}(\mathbf{\tilde{K}}^{\mathrm{d}}_i))
\end{equation}
using \gls{mhsa}~\cite{vaswani2017attention} followed by a \gls{ffn}.
The updated kernels $\mathbf{K}^{\mathrm{p}}_i$ and $\mathbf{K}^{\mathrm{d}}_i$ are fed through separate \glspl{ffn} 
\begin{equation}
    \mathbf{M}^{\mathrm{p}}_i = \mathrm{FFN}^{\mathrm{p}}_{\mathbf{M}}(\mathbf{K}^{\mathrm{p}}_i) \ast \mathbf{F}^{\mathrm{p}},\quad \mathbf{M}^{\mathrm{d}}_i = \mathrm{FFN}^{\mathrm{d}}_{\mathbf{M}}(\mathbf{K}^{\mathrm{d}}_i) \ast \mathbf{F}^{\mathrm{d}},\quad \mathbf{p}_i = \mathrm{FFN}_{\mathbf{p}}(\mathbf{K}^{\mathrm{p}}_i)
\end{equation}
to generate refined panoptic mask predictions $\mathbf{M}^{\mathrm{p}}_i$, self-supervised depth mask predictions $\mathbf{M}^{\mathrm{d}}_i$, and a class probability prediction $\mathbf{p}_i$.

\subsection{Self-supervised Depth Predictor}
\label{sec:depth_pred}
\begin{figure}[tb]
  \centering
  \begin{subfigure}{0.49\linewidth}
    \centering
    \includegraphics[width=\linewidth]{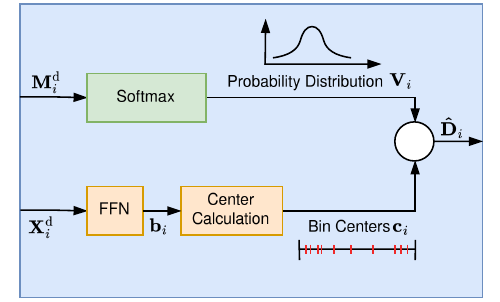}
    \caption{Self-supervised depth predictor}
    \label{fig:depth_pred}
  \end{subfigure}
  \hfill
  \begin{subfigure}{0.49\linewidth}
    \centering
    \includegraphics[width=\linewidth]{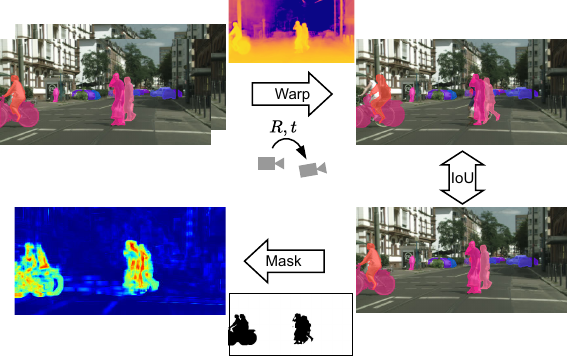}
    \caption{Panoptic-guided motion masking}
    \label{fig:motion_mask}
  \end{subfigure}
  \caption{Visualization of our self-supervised depth predictor module in (a) and our panoptic-guided motion masking in (b).}
\end{figure}

We introduce a self-supervised depth predictor module that efficiently generates an inverse depth prediction $\mathbf{\hat{D}}_i$ based on depth mask predictions $\mathbf{M}^{\mathrm{d}}_i$ and depth group features $\mathbf{X}^{\mathrm{d}}_i$.
Inspired by recent approaches~\cite{bhat2021adabins, wang2023sqldepth}, we reformulate depth estimation as a classification-regression problem and predict $N$ depth bins $\mathbf{b}_i\in\mathbb{R}^{B\times N}$ using a combination of depth planes and bin centers.
Unlike recent methods, we directly use the encoded panoptic information present in our model to predict depth bins at the panoptic mask level.
\Cref{fig:depth_pred} gives an overview of our self-supervised depth predictor module.
The depth group features $\mathbf{X}^{\mathrm{d}}_i$ are calculated based on the depth feature map $\mathbf{F}^{\mathrm{d}}$ and the normalized binary panoptic mask predictions $\mathbf{S}^{\mathrm{p}}_{i-1}$ in the \gls{gfa} stage of the linked kernel update heads.
Thus, they encode object features at a panoptic mask level, which we leverage to predict depth bins at the panoptic mask level.
To this end, we first reshape the depth group features to shape $B\times N C$ and then feed them through a simple \gls{ffn}
\begin{equation}
   \mathbf{b}_i = \mathrm{FFN}^{\mathrm{d}}_{\mathbf{b}}(\mathbf{X}^{\mathrm{d}}_i).
\end{equation}
Following~\cite{bhat2021adabins}, we convert the predicted bins into bin centers $\mathbf{c}_i\in\mathbb{R}^{B\times N}$
\begin{equation}
   \mathbf{c}_i = d_{\mathrm{min}} + (d_{\mathrm{max}} - d_{\mathrm{min}})\left( \frac{\mathbf{b}_i}{2} + \sum_{j=1}^{i-1}\mathbf{b}_j \right)
\end{equation}
using the minimum depth $d_{\mathrm{min}}$ and maximum depth $d_{\mathrm{max}}$.
The depth mask predictions $\mathbf{M}^{\mathrm{d}}_i$ are converted into a probability distribution $\mathbf{V}_i\in\mathbb{R}^{B\times N \times H' \times W'}$ over the predicted depth bins by performing a softmax operation over the mask predictions
\begin{equation}
   \mathbf{V}_i = \mathrm{softmax}(\mathbf{M}^{\mathrm{d}}_i).
\end{equation}
Finally, we use the linear combination of depth centers and softmax scores to calculate the inverse depth prediction $\mathbf{\hat{D}}_i$.
Using the bin-wise formulation at the panoptic mask level has several advantages.
First, it uses the panoptic information present in our model efficiently without relying on additional modules such as the Self-Query Layer used in~\cite{wang2023sqldepth}, which is computationally expensive when processing high-resolution images ($\mathcal{O}(H'\times W')$).
Second, the resulting depth maps are aligned with the panoptic prediction and thus encompass fine-grained details.

\subsection{Post-processing}
\label{sec:post_proc}
We use the post-processing module introduced in RT-K-Net~\cite{schoen2023rt} to generate panoptic segmentation predictions based on the updated panoptic masks $\mathbf{M}^{\mathrm{p}}_4$ and class probability prediction $\mathbf{p}_4$.
For depth estimation, we use the post-processing introduced in MGNet~\cite{schoen2021mgnet}, where we use the panoptic road prediction as ground mask to calculate the depth scale factor based on the relation between the estimated median camera height for ground pixels and the calibrated camera height.

\clearpage

\subsection{Optimization}
\label{sec:optimization}
For panoptic segmentation, we follow RT-K-Net and use bipartite matching to match mask predictions with ground-truth segments.
The matched segments are used together with auxiliary outputs to calculate the panoptic loss
\begin{equation}
    \mathcal{L}^{\mathrm{p}} = \lambda_{\mathrm{mask}}\mathcal{L}^{\mathrm{mask}} + \lambda_{\mathrm{dice}}\mathcal{L}^{\mathrm{dice}} + \lambda_{\mathrm{cls}}\mathcal{L}^{\mathrm{cls}} + \lambda_{\mathrm{rank}}\mathcal{L}^{\mathrm{rank}} + \lambda_{\mathrm{seg}}\mathcal{L}^{\mathrm{seg}} + \lambda_{\mathrm{inst}}\mathcal{L}^{\mathrm{inst}}
\end{equation}
with $\mathcal{L}^{\mathrm{mask}}$ being the binary cross-entropy loss, $\mathcal{L}^{\mathrm{dice}}$ being the dice loss~\cite{sudre2017generalised}, $\mathcal{L}^{\mathrm{cls}}$ being the focal loss for classification~\cite{lin2017focal}, $\mathcal{L}^{\mathrm{rank}}$ being the mask-id cross-entropy loss, $\mathcal{L}^{\mathrm{seg}}$ being the cross-entropy loss, and $\mathcal{L}^{\mathrm{inst}}$ being the instance discrimination loss~\cite{yu2022cmt}.
The weight factors $\lambda_{\mathrm{mask}}, \lambda_{\mathrm{dice}}, \lambda_{\mathrm{cls}}, \lambda_{\mathrm{rank}}, \lambda_{\mathrm{seg}}$ and $\lambda_{\mathrm{inst}}$ are hyperparameters and used to balance the different loss terms.
We follow common practice~\cite{godard2019digging, schoen2021mgnet} and use a combination of photometric loss and smoothness loss for self-supervised depth estimation
\begin{equation}
    \mathcal{L}^{\mathrm{d}} = \lambda_{\mathrm{phot}}\mathcal{L}^{\mathrm{phot}} + \lambda_{\mathrm{smooth}}\mathcal{L}^{\mathrm{smooth}}.
\end{equation}
A self-supervised depth predictor module is added to each linked kernel update head to predict four inverse depth predictions $\mathbf{\hat{D}}_{1...4}$, which are used for loss calculation.
We use a gradient-based smoothness loss~\cite{godard2017unsupervised}
\begin{equation}
    \mathcal{L}^{\mathrm{smooth}} = \sum_{i=1}^4|\delta_x\mathbf{\hat{D}}_i^*|\exp(-|\delta_x{\mathbf{I}}^{(t)}|)\\ 
+ |\delta_y\mathbf{\hat{D}}_i^*|\exp(-|\delta_y{\mathbf{I}}^{(t)}|)
\end{equation}
with predicted mean-normalized inverse depth $\mathbf{\hat{D}}_i^*$ and reference image $\mathbf{I}^{(t)}$.
For photometric loss calculation, we use a separate Pose-CNN~\cite{wang2023sqldepth} to predict the 6~DOF relative camera pose between adjacent frames.
For each depth prediction $\mathbf{D}_i = 1 / \mathbf{\hat{D}}_i$, the predicted poses are used to warp the context frames $\mathbf{I}_c$ into the reference frame $\mathbf{I}^{(t)}$.
For each warped image $\mathbf{\hat{I}}^{(t)}_i\in\{\mathbf{\hat{I}}^{(t-1)\rightarrow (t)}_i, \mathbf{\hat{I}}^{(t+1)\rightarrow (t)}_i\}$, the photometric loss is calculated according to
\begin{equation}
\mathcal{L}^{\mathrm{phot}}(\mathbf{I}^{(t)}, \mathbf{\hat{I}}^{(t)}_i) = \alpha \frac{1-\text{SSIM}(\mathbf{I}^{(t)}, \mathbf{\hat{I}}^{(t)}_i)}{2} + (1-\alpha) |\mathbf{I}^{(t)}-\mathbf{\hat{I}}^{(t)}_i|
\end{equation}
using the \gls{ssim}~\cite{wang2004image}.
Following common practice~\cite{godard2019digging}, we set $\alpha=0.85$.
Furthermore, we use the minimum reprojection error between warped and unwarped frames to mask out static pixels and the auto masking strategy introduced in~\cite{godard2019digging}.
Finally, we calculate a panoptic-guided motion mask as described in the following section to mask out pixels that potentially belong to dynamic objects
\begin{equation}
\mathcal{L}^{\mathrm{phot}} = \sum_{i=1}^4\mathcal{L}^{\mathrm{phot}}(\mathbf{I}^{(t)}, \mathbf{\hat{I}}^{(t)}_i)\odot \mathbf{M}_{\mathrm{dyn}}.
\end{equation}
The final loss is calculated as a combination of panoptic segmentation and depth loss terms
\begin{equation}
    \mathcal{L} = \mathcal{L}^{\mathrm{p}} + \lambda_{\mathrm{depth}}\mathcal{L}^{\mathrm{d}}
\end{equation}
with weight factor $\lambda_{\mathrm{depth}}$ to balance both loss terms.

\subsection{Panoptic-guided Motion Mask Calculation}
\label{sec:motion_mask}
To tackle the problem of dynamic objects, which break the static scene assumption in the self-supervised training objective, we introduce a panoptic-guided motion mask to mask out potential dynamic objects from the photometric loss calculation.
\Cref{fig:motion_mask} gives an overview of the motion mask calculation.
First, forward passes through the model are performed for the context images $\mathbf{I}_c$ to receive panoptic predictions for both context images $\mathbf{P}_c\in\{\mathbf{P}^{(t-1)}, \mathbf{P}^{(t+1)}\}$.
We employ a tracking module~\cite{stolle2023unified} to get consistent instance IDs for the three panoptic predictions $\mathbf{\hat{P}}^{(t-1)}, \mathbf{\hat{P}}^{(t)}$ and $\mathbf{\hat{P}}^{(t+1)}$.
The advantage of this module is two-fold:
First, it combines appearance-based matching and position-based matching for instance tracking.
Second, the module does not require annotations with consistent instance IDs, which are costly to acquire for real-world applications.
The tracked panoptic segmentation predictions are then warped into the reference frame using the predicted camera poses and depth prediction $\mathbf{D}$.

Finally, similar to~\cite{petrovai2023monodvps}, the \glspl{iou} between the warped and reference panoptic predictions are calculated for all segments corresponding to potential dynamic objects such as cars or pedestrians.
Segments with a \gls{iou} below a threshold $T$ are considered part of a dynamic object and thus added to the motion mask $\mathbf{M}_{\mathrm{dyn}}$.
In contrast to~\cite{petrovai2023monodvps}, we achieve the best results with a fixed threshold of $T=0.5$.
Furthermore, since we use predictions instead of ground truth segmentation maps, initial predictions contain much noise.
Thus, we turn off the motion mask calculation for the first $k=1000$ training iterations until the model predicts reasonable object instances.

\section{Experiments}
\label{sec:experiments}
\subsection{Datasets}
\label{sec:datasets}
We provide experimental results on two popular autonomous driving benchmarks, Cityscapes~\cite{cordts2016cityscapes} and KITTI~\cite{geiger2013vision}.
The Cityscapes dataset provides video sequences recorded in 50 cities, primarily in Germany.
It contains 5\,000 images annotated with fine-grained panoptic labels for 30 classes, of which 19 are used during evaluation.
The dataset is split into 2\,975 images for training, 500 for validation, and 1\,525 for testing.
Depth annotations are given as pre-computed disparity maps from SGM~\cite{hirschmuller2008stereo} for the 5\,000 fine annotated images, which are only used for depth evaluation.
For KITTI, we use the KITTI Eigen split introduced in~\cite{eigen2014depth} with pre-processing from Zhou~\etal~\cite{zhan2018unsupervised} to remove static frames.
This leads to an image split of 39\,810 images for training and 4424 for validation.
Since the KITTI Eigen split does not provide panoptic annotations, we generate panoptic pseudo-labels using our best-performing model on Cityscapes and train our method using only the generated pseudo-labels.
We evaluate depth estimation performance on KITTI on the 652 improved ground truth depth maps introduced in~\cite{uhrig2017sparsity}.

\subsection{Implementation Details}
\label{sec:impl_details}
All models are trained on 4 NVIDIA A6000 GPUs; inference times are measured on a single NVIDIA Titan RTX GPU without TensorRT~\cite{TensorRT} optimization.
The AdamW optimizer~\cite{loshchilov2019decoupled} is used with the \textit{poly} learning rate scheduler~\cite{liu2016parsenet}, and a base learning rate of 0.0001.
We use a total batch size of 8, weight decay of 0.05, and train for 60k iterations.
For data augmentation, we use random image scaling between 0.5 and 2.1, random cropping with the instance crop method introduced by RT-K-Net~\cite{schoen2023rt}, random image color and brightness adjustment, and random left-right flip.
For Cityscapes, we use a crop size of $1024 \times 2048$ pixel; for KITTI, we resize images to a fixed size of $384 \times 1280$ pixel.
For panoptic loss weights, we follow RT-K-Net settings and set $\lambda_{mask}=1.0, \lambda_{dice}=4.0, \lambda_{rank}=0.1, \lambda_{cls}=2.0, \lambda_{seg}=1.0$, and $\lambda_{inst}=1.0$.
For depth loss weights, we set $\lambda_{phot}=1.0$, $\lambda_{smooth}=0.001$, and $\lambda_{depth}=25.0$ to balance panoptic and depth loss terms. 
We investigate the effect of different values for $\lambda_{depth}$ in~\cref{sec:depth_loss}.
We use $N=100$ and $C=256$.
Following recent combined methods~\cite{yuan2022polyphonicformer, zhang2023unidaformer}, we initialize our model with weights pre-trained on the Mapillary Vistas dataset~\cite{neuhold2017mapillary} for panoptic segmentation unless stated otherwise.
For Cityscapes, \gls{pq}~\cite{kirillov2019panoptic1} is the primary metric for panoptic segmentation, which can also be calculated for stuff (PQ$_{\text{st}}$) and thing (PQ$_{\text{th}}$) classes separately, while RMSE is the primary metric for self-supervised depth estimation.
For KITTI, we report standard depth metrics as used in~\cite{eigen2014depth, xue2020hierarchical, guizilini20203d, watson2021temporal, zhao2022monovit, wang2023sqldepth}.
The reader is referred to the respective publication~\cite{kirillov2019panoptic1, eigen2014depth} for details on the respective metrics.

\subsection{Experimental Results}
\label{sec:results}

\subsubsection{Cityscapes.}
\Cref{tab:cityscapes} shows results on the Cityscapes dataset compared to state-of-the-art methods. 
\Cref{tab:panoptic} shows the comparison with real-time panoptic segmentation methods while \Cref{tab:depth_cityscapes} shows the comparison with depth estimation methods.
For a fair comparison, we report both results with and without Mapillary Vistas pre-training and inference times with and without automatic-mixed precision; denoted FP16 and FP32, respectively.
For panoptic segmentation, MGNiceNet outperforms all other real-time capable methods regarding panoptic quality.
Compared to our previous state-of-the-art approach RT-K-Net~\cite{schoen2023rt}, PQ is improved by 1.5\% without using additional data and 3.8 \% when using Mapillary Vistas pre-training.
Regarding runtime, MGNiceNet has a higher runtime of $42~\si{ms}$ than RT-K-Net's $32~\si{ms}$. 
Still, the computational overhead is marginal, given that MGNiceNet performs the additional task of self-supervised depth estimation.
Compared to MGNet~\cite{schoen2021mgnet}, MGNiceNet has a slightly higher runtime, but the performance gain in terms of accuracy is large enough to justify the higher runtime. 
A comparison of MGNiceNet to non-real-time panoptic segmentation methods, such as~\cite{yuan2022polyphonicformer}, is carried out in the paper's supplementary.
For depth estimation, MGNiceNet outperforms all other methods regarding the RMSE metric when using Mapillary Vistas pre-training.
Without the pre-training, MGNiceNet still outperforms MGNet by $1~\si{m}$ RMSE and closes the gap to supervised methods~\cite{xu2018pad, zhang2018joint, wang2020sdc, gao2022panopticdepth}, which are not focused on real-time application, significantly.

\begin{table}[!t]
\caption{\textbf{Main results on the Cityscapes dataset.} A comparison to real-time panoptic segmentation methods on the Cityscapes validation set is shown in (a), while a comparison to depth estimation methods on the Cityscapes test set is shown in (b). Methods marked with\,\dag\, use additional data,~\eg, Mapillary Vistas~\cite{neuhold2017mapillary} pre-training.}
\label{tab:cityscapes}
\captionsetup{labelfont=normalsize,textfont=normalsize,position=top}
\centering
\scalebox{0.69}{\subfloat[Real-time panoptic segmentation]{%
    \label{tab:panoptic}
    \begin{tabularx}{10.5cm}{l|c|c|c|c|Y|Y}
        \toprule
        \multirow{2}{*}{Method} & \multirow{2}{*}{PQ $\uparrow$} & \multirow{2}{*}{PQ$_{\text{th}}$ $\uparrow$} & \multirow{2}{*}{PQ$_{\text{st}}$ $\uparrow$} & \multirow{2}{*}{GPU} & \multicolumn{2}{c}{Runtime [ms] $\downarrow$} \\
        & & & & & FP16 & FP32 \\
        \midrule
        MGNet~\cite{schoen2021mgnet}\dag & 55.7 & 45.3 & 63.1 & Titan RTX & \underline{36} & - \\
        Panoptic-DeepLab~\cite{chen2020scaling} & 58.4 & - & - & V100 & - & \textbf{63} \\
        Hou~\etal~\cite{hou2020real} & 58.8 & 52.1 & 63.7 & V100 & - & \underline{99} \\
        RT-K-Net~\cite{schoen2023rt} & 60.2 & 51.5 & 66.5 & Titan RTX & \textbf{32} & - \\
        \midrule
        \textbf{Ours} & \underline{61.7} & \underline{54.6} & \underline{66.8} & Titan RTX & 42 & \textbf{63} \\
        \textbf{Ours}\dag & \textbf{64.0} & \textbf{56.3} & \textbf{69.5} & Titan RTX & 42 & \textbf{63} \\
        \bottomrule
    \end{tabularx}
}}
\hfill
\scalebox{0.69}{\subfloat[Depth estimation]{%
    \label{tab:depth_cityscapes}
    \begin{tabularx}{6.0cm}{l|c|c} 
        \toprule
        Method & RMSE $\downarrow\,$ & GT Depth \\
        \midrule
        MGNet~\cite{schoen2021mgnet}\dag & 8.3 & \\
        Pad-Net~\cite{xu2018pad} & 7.12 & \checkmark \\
        Zhang~\etal~\cite{zhang2018joint} & 7.10 & \checkmark \\
        SDC-Depth~\cite{wang2020sdc} & 6.92 & \checkmark \\
        PanopticDepth~\cite{gao2022panopticdepth}\dag & \underline{6.69} & \checkmark \\
        \midrule
        \textbf{Ours} & 7.3 & \\
        \textbf{Ours}\dag & \textbf{6.63} & \\
        \bottomrule
    \end{tabularx}
}}
\end{table}

\subsubsection{KITTI Eigen.}
\label{sec:results_kitti}
\Cref{tab:kitti} shows results on the KITTI Eigen split compared to state-of-the-art methods in self-supervised depth estimation.
MGNiceNet outperforms most methods~\cite{godard2019digging, kumar2021syndistnet, schoen2021mgnet} by a large margin and even outperforms PackNet-SfM~\cite{guizilini20203d}, which uses a much heavier architecture.
Only the state-of-the-art approaches MonoViT~\cite{zhao2022monovit} and SQLdepth~\cite{wang2023sqldepth} slightly outperform \mbox{MGNiceNet} in most metrics.
We reason that this is primarily because our method is designed with a focus on low latency, while the aforementioned approaches rely on computationally expensive layers such as the Self Query Layer used in~\cite{wang2023sqldepth}.
This becomes apparent when comparing inference times.
Since most methods do not provide inference times in the paper, we use the author's official codebases, if available, to get comparable inference times.
MGNiceNet has an inference time of $21~\si{ms}$ when using automatic mixed-precision, which is around 75\% faster than SQLdepth and 50 \% faster than MonoViT.
Thus, MGNiceNet provides a much better accuracy-speed trade-off than~\cite{zhao2022monovit, wang2023sqldepth}.
Furthermore, MGNiceNet tackles the additional task of panoptic segmentation.
Compared to MGNet, we see similar results as on Cityscapes, with MGNet slightly outperforming MGNiceNet in terms of runtime.
We show qualitative results on Cityscapes and KITTI and direct qualitative comparisons in the paper's supplementary.

\begin{table}[t!]
\caption{\textbf{Main results on the KITTI Eigen split.} Depth estimation is evaluated for a distance up to 80~\si{m} using the improved ground truth maps from~\cite{uhrig2017sparsity}. Runtimes marked with\,*\, are calculated based on official code releases.}
\label{tab:kitti}
\centering
\scalebox{0.69}{\begin{tabularx}{17cm}{l|c|c|c|c|c|c|c|c|Y|Y}
\toprule
\multirow{2}{*}{Method} & \multirow{2}{*}{DS} & \multirow{2}{*}{Resolution} & \multirow{2}{*}{Abs Rel $\downarrow$} & \multirow{2}{*}{RMSE $\downarrow$} & \multirow{2}{*}{$\delta<1.25 \uparrow$} & \multirow{2}{*}{$\delta<1.25^2 \uparrow$} & \multirow{2}{*}{$\delta<1.25^3 \uparrow$} & \multirow{2}{*}{GPU} &  \multicolumn{2}{c}{Runtime [ms] $\downarrow$} \\
& & & & & & & & & FP16 & FP32 \\
\midrule
Monodepth2~\cite{godard2019digging} & K & $640\times192$ & 0.090 & 3.942 & 0.914 & 0.983 & 0.995 & - & - & - \\
MGNet~\cite{schoen2021mgnet} & CS+K & $1280\times384$ & 0.095 & 3.761 & 0.902 & 0.979 & 0.992 & Titan RTX & \textbf{13}* & \underline{34}* \\
SynDistNet~\cite{kumar2021syndistnet} & K & $640\times192$ & 0.076 & 3.406 & 0.931 & 0.988 & 0.996 & - & - & - \\
PackNet-SfM~\cite{guizilini20203d} & CS+K & $1280\times384$ & 0.071 & 3.153 & 0.944 & 0.990 & 0.997 &  V100 & - & 60 \\
MonoViT~\cite{zhao2022monovit} & K &$1280\times384$ & \underline{0.067} & \underline{3.108} & \underline{0.950} & \underline{0.992} & \textbf{0.998} & Titan RTX & 41* & 56* \\
SQLdepth~\cite{wang2023sqldepth} & K &$1024\times320$ & \textbf{0.058} & \textbf{2.925} & \textbf{0.962} & \textbf{0.993} & \textbf{0.998} & Titan RTX & 99* & 107* \\
\midrule
\textbf{Ours} & CS+K & $1280\times384$ & 0.069 & 3.143 & 0.949 & \underline{0.992} & \textbf{0.998} & Titan RTX & \underline{21} & \textbf{28} \\
\bottomrule
\end{tabularx}}
\end{table}

\subsection{Ablation Studies}
\label{sec:ablation}
\begin{table}[!t]
    \captionsetup{labelfont=small,textfont=small,position=top}
	\label{tab:ablation}
	\caption{\textbf{Ablation studies for MGNiceNet.} All experiments are conducted on the Cityscapes validation dataset.}
	\centering
    \scalebox{0.85}{\subfloat[Multi-task learning vs single-task performance]{%
        \label{tab:mtl}
        \begin{tabularx}{8.3cm}{l|c|c|c|c|c} 
            \toprule
            Method & Depth & Panoptic & PQ $\uparrow$  & RMSE $\downarrow$ & Runtime [ms] $\downarrow$ \\
            \midrule
            Depth & \checkmark & -  &  - & 7.9 & 27 \\ 
            Panoptic & - & \checkmark & 64.0 & - & 32 \\
            \midrule
            Ours & \checkmark & \checkmark & 64.0 & 7.1 & 42 \\
            \bottomrule
        \end{tabularx}
    }}
    \hfill
    \scalebox{0.85}{\subfloat[Loss balancing]{%
        \label{tab:depth_loss}
        \begin{tabularx}{3.1cm}{l|c|c} 
            \toprule
            $\lambda_{depth}$ & PQ $\uparrow$ & RMSE $\downarrow$ \\
            \midrule
                5.0 & 64.0 & 7.7 \\
                15.0 & 63.9 & 7.5 \\
                25.0 & 64.0 & 7.1 \\
                30.0 & 63.3 & 7.1 \\
            \bottomrule
        \end{tabularx}
    }}
    \hfill
    \vspace{1mm}
    \scalebox{0.85}{\subfloat[Motion masking]{%
        \label{tab:extension}
        \begin{tabular}{l|c|c|c} 
            \toprule
            Method & Motion masking & PQ $\uparrow$  & RMSE $\downarrow$ \\
            \midrule
            \multirow{2}{*}{Ours} & -  & 63.6 & 7.8 \\ 
            & \checkmark & 64.0 & 7.1 \\
            \bottomrule
        \end{tabular}
    }}
    \hfill
    \scalebox{0.85}{\subfloat[Depth predictor choice]{%
        \label{tab:depth_pred}
		\begin{tabularx}{6.6cm}{l|c|c|c} 
			\toprule
			Method & PQ $\uparrow$ & RMSE $\downarrow$ & Runtime [ms] $\downarrow$ \\
			\midrule
            Instance-Wise & 63.3 & 7.3 & 45 \\
            Ours & 64.0 & 7.1 & 42 \\
			\bottomrule
		\end{tabularx}
    }}
\end{table}

\subsubsection{Multi-Task learning vs Single-Task Performance.}
We compare our multi-task approach to single-task baselines by training MGNiceNet without depth estimation or panoptic segmentation, respectively.
As seen in \cref{tab:mtl}, our multi-task approach has the same performance in terms of panoptic quality but improves depth performance compared to the single-task baselines.
Furthermore, our method only adds marginal runtime while handling both tasks.
We argue that this improved depth performance results from our contributions, namely our unified MGNiceNet architecture, the self-supervised depth predictor, and our panoptic-guided motion masking strategy.

\subsubsection{Loss Balancing.}
\label{sec:depth_loss}
We compare models trained with different depth loss weights \mbox{$\lambda_{depth}\in\{5.0, 15.0, 25.0, 30.0\}$} in \cref{tab:depth_loss}.
As expected, depth estimation performance increases for higher values of $\lambda_{depth}$ while panoptic segmentation performance decreases.
Therefore, we choose $\lambda_{depth} = 25.0$, which achieves the best performance in both tasks.

\subsubsection{Motion Masking.}
\begin{figure}[t]
\centering
\subfloat[Input image]{%
    \includegraphics[width=0.3\linewidth]{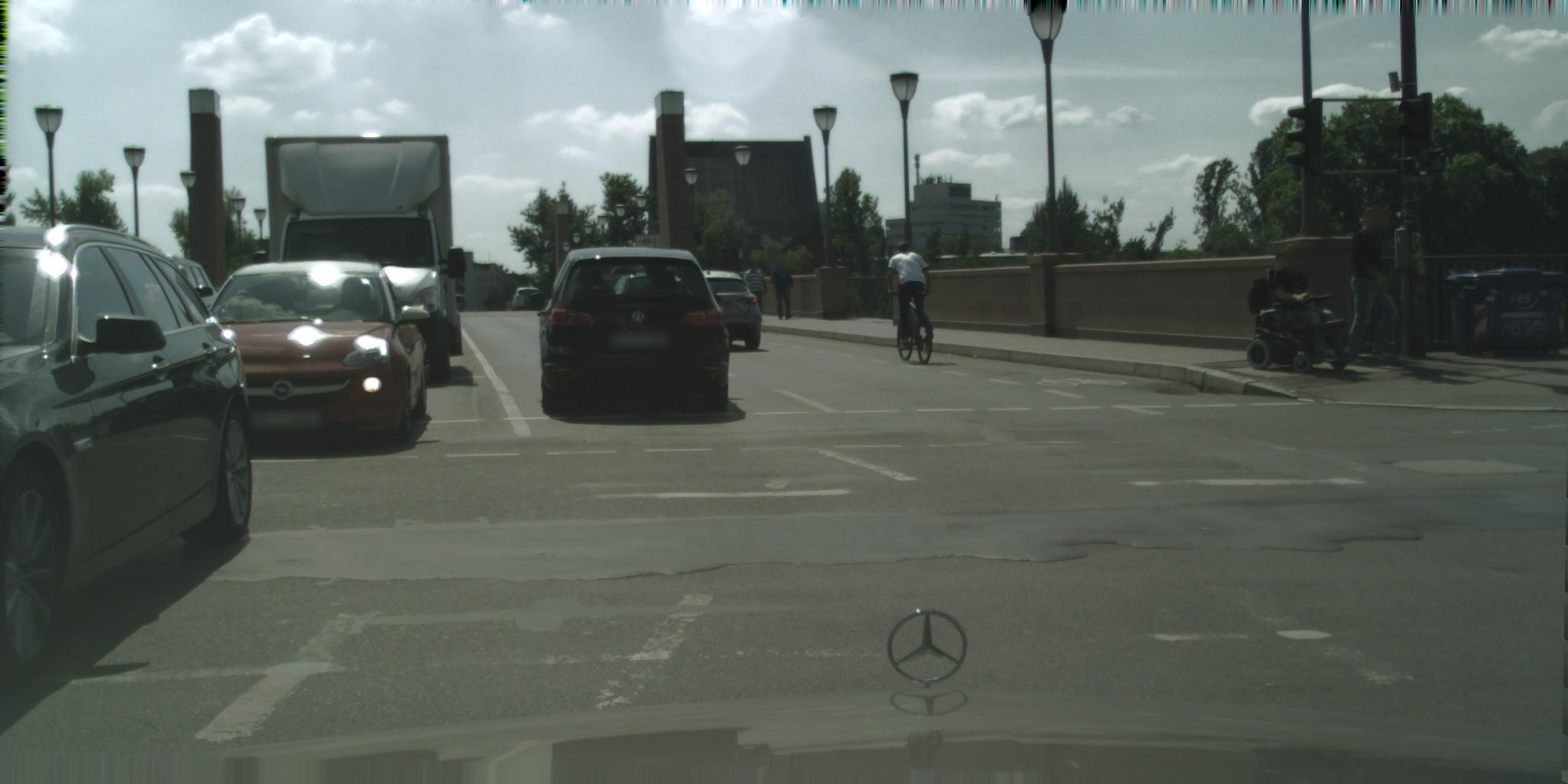}%
}
\hfill
\subfloat[Without motion mask]{%
    \includegraphics[width=0.3\linewidth]{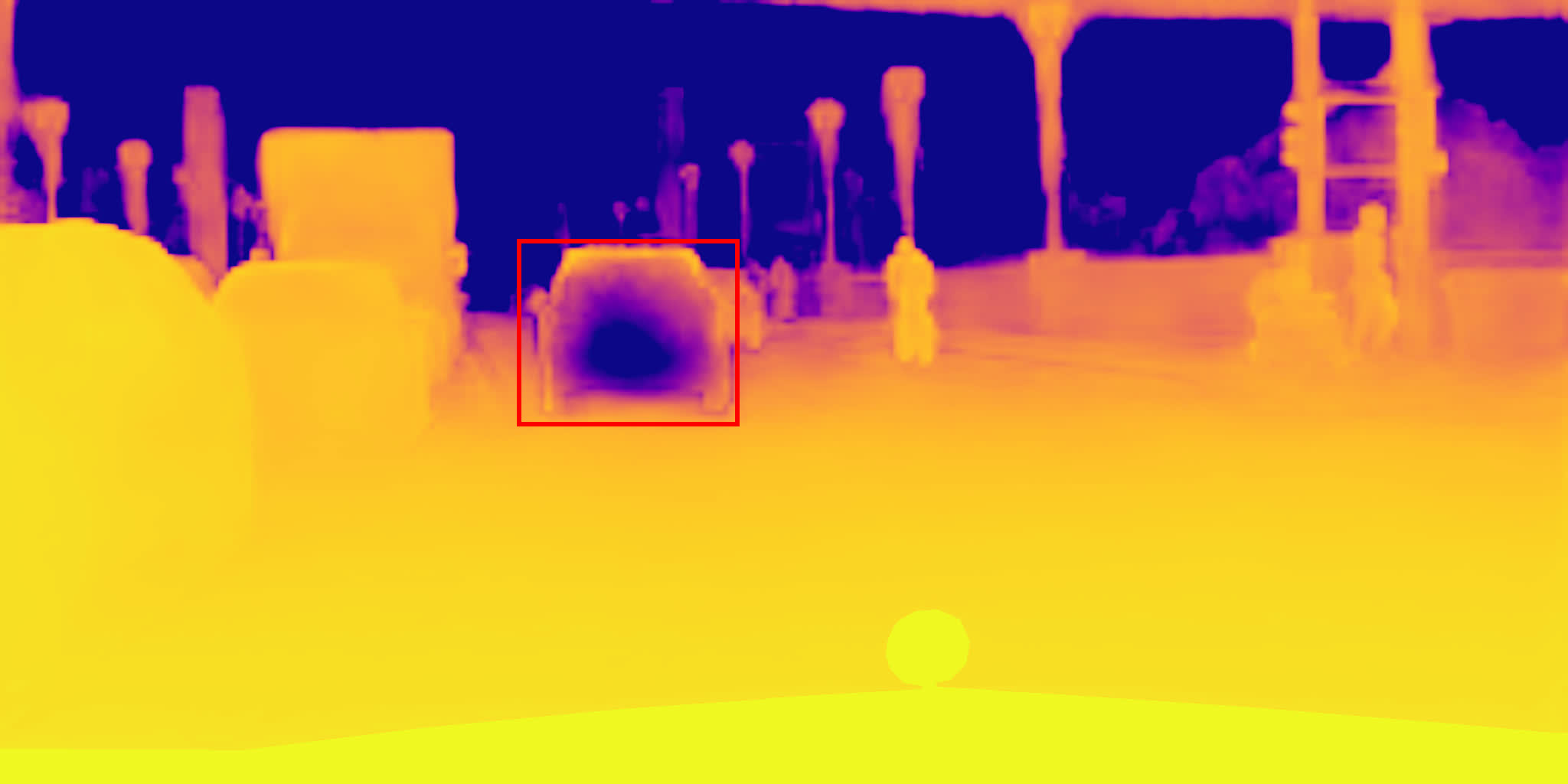}%
}
\hfill
\subfloat[With motion mask]{%
    \includegraphics[width=0.3\linewidth]{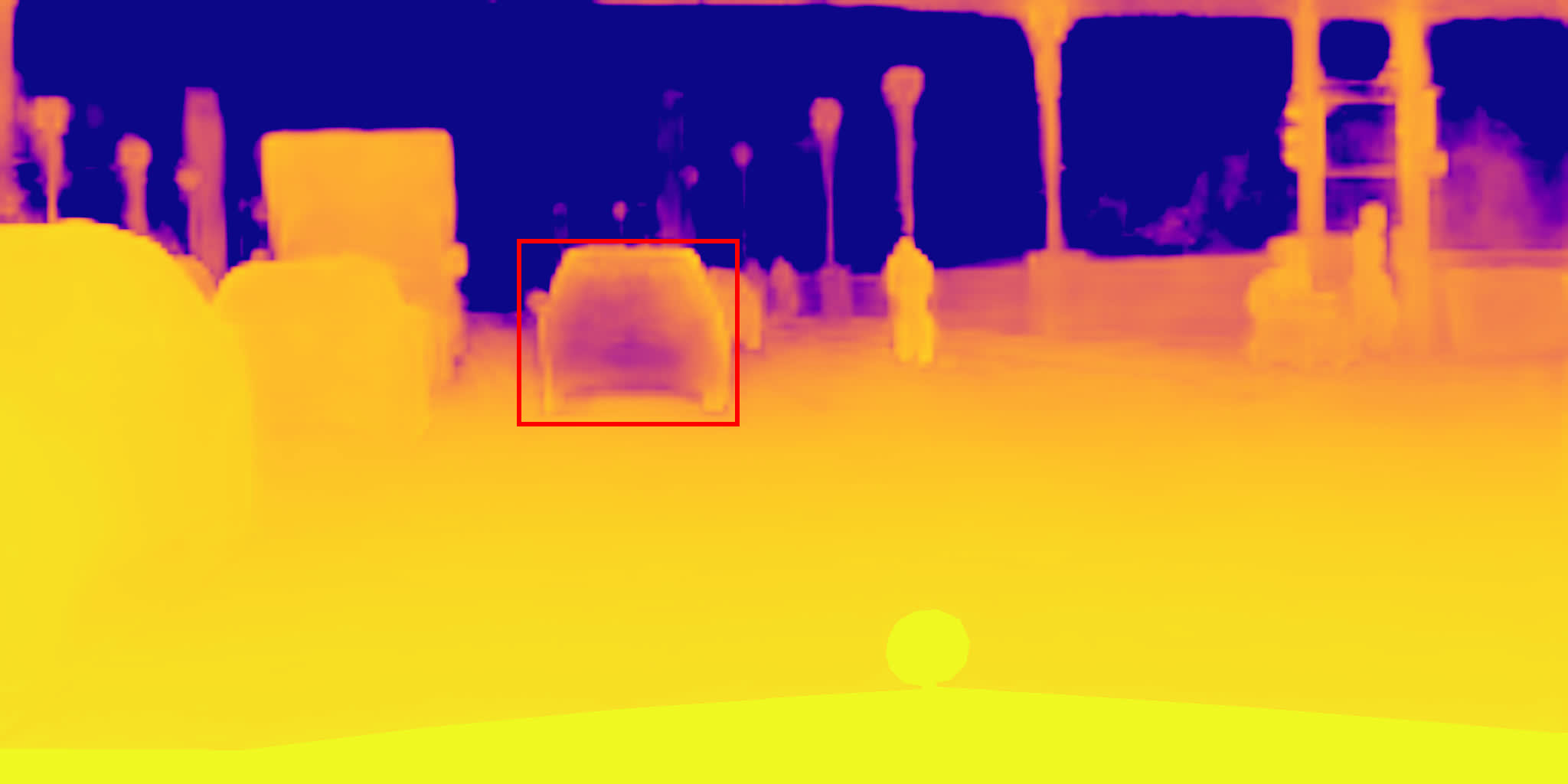}%
}
\caption{Effect of our panoptic-guided motion masking. While the model predicts holes of infinite depth on pixels corresponding to the car driving with a similar velocity as the ego vehicle, the effect is reduced significantly when using our panoptic-guided motion masking method during training.}
\label{fig:motion_mask_result}
\end{figure}

We investigate the effects of our panoptic-guided motion masking as described in~\cref{sec:motion_mask} by training a model without and with our proposed motion masking approach.
\cref{tab:extension} shows that adding our panoptic-guided motion masking significantly improves depth estimation performance while panoptic quality remains nearly the same.
This can also be seen in~\cref{fig:motion_mask_result}, which shows depth predictions of a model trained with and without our motion masking method.
We want to especially highlight that the model trained without our panoptic-guided motion masking still uses MonoDepth2's~\cite{godard2019digging} auto-masking strategy, which fails for datasets with a high number of dynamic objects such as Cityscapes.

\subsubsection{Depth Predictor Choice.}
We compare our self-supervised depth predictor as described in~\cref{sec:depth_pred} with an instance-wise depth predictor.
The instance-wise depth predictor uses depth mask predictions $\mathbf{M}^{\mathrm{d}}_4$ and binarized mask predictions $\mathbf{S}^{\mathrm{p}}_4$ to generate a depth prediction by gathering all pixels in $\mathbf{M}^{\mathrm{d}}_4$ where $\mathbf{S}^{\mathrm{p}}_4 = 1$.
This is a similar strategy to PolyphonicFormer~\cite{yuan2022polyphonicformer}, which trains instance-wise depth predictions in a supervised way.
\Cref{tab:depth_pred} shows that our self-supervised depth predictor outperforms the instance-wised predictor in both metrics.
Interestingly, the instance-wise predictor also negatively affects panoptic segmentation performance.
We argue this is due to our method's lack of instance-wise supervision with ground truth depth labels.

\subsubsection{Limitations and Future Work.}
Our method is optimized to process high-resolution images at a fast inference speed.
As described in~\cref{sec:results_kitti}, recent state-of-the-art methods still slightly outperform our approach.
In future work, we aim to improve performance further, especially when dealing with lower-resolution input images, such as images from the KITTI dataset.
\section{Conclusion}
\label{sec:conclusion}
This work introduces MGNiceNet, a unified monocular geometric scene understanding approach.
We extend our real-time panoptic segmentation method RT-K-Net to tackle the additional task of self-supervised depth estimation using linked kernel update heads and a lightweight depth predictor.
Our self-supervised depth predictor uses a bin-wise depth representation at the panoptic mask level to predict high-quality depth maps that align well with panoptic predictions.
We tackle the dynamic object problem in self-supervised depth estimation using a novel panoptic-guided motion masking method.
We show our method's effectiveness and limitations in extensive experiments on the Cityscapes and KITTI datasets.
Our method provides an excellent speed-accuracy trade-off and closes the gap to more computationally demanding methods.

\begin{credits}
\subsubsection{\ackname} This research is accomplished within the project ``AUTOtech.agil'' (FKZ 01IS22088W). We acknowledge the financial support for the project by the Federal Ministry of Education and Research of Germany (BMBF).
\end{credits}

\bibliographystyle{splncs04}
\bibliography{main}
\end{document}


\title{MGNiceNet: Unified Monocular Geometric Scene Understanding\\ - \\Supplementary Material} 
\titlerunning{MGNiceNet: Unified Monocular Geometric Scene Understanding}

\author{Markus Schön\orcidlink{0009-0006-6029-9341} \and
Michael Buchholz\orcidlink{0000-0001-5973-0794} \and
Klaus Dietmayer\orcidlink{0000-0002-1651-014X}%
}

\authorrunning{M. Schön~\etal}

\institute{Institute of Measurement, Control, and Microtechnology, Ulm University, Germany
\email{\{markus.schoen,michael.buchholz,klaus.dietmayer\}@uni-ulm.de}}

\maketitle

\appendix

\section{Comparison to SOTA Panoptic Segmentation Methods}
\label{sec:supp_ablation}
We compare our method to non-real-time state-of-the-art panoptic segmentation methods on the Cityscapes~\cite{cordts2016cityscapes} validation dataset in \cref{tab:cityscapes_sota_comp}.
\mbox{MGNiceNet} focuses on low latency for real-time applications such as autonomous driving perception systems.
Thus, heavier architectures that do not consider this constraint naturally outperform our approach in terms of panoptic quality.
However, \cref{tab:cityscapes_sota_comp} shows that our approach can significantly close the gap to heavier approaches compared to the current real-time state-of-the-art method RT-K-Net~\cite{schoen2023rt}.
Compared to Axial-DeepLab-XL~\cite{wang2020axial}, which is the best-performing approach that neither uses Mapillary Vistas~\cite{neuhold2017mapillary} nor ImageNet-22K~\cite{deng2009imagenet} pre-training, our approach can close the gap from $4.2\%$~PQ to $2.7\%$~PQ, a significant improvement of approximately 35\%.
Compared to the current state-of-the-art approach OneFormer~\cite{jain2023oneformer} with ConvNeXt-L~\cite{liu2022convnet} backbone, which uses both Mapillary Vistas and ImageNet-22K pre-training, our approach is outperformed by $6.1\%$~PQ.
We argue that this large gap is due to ImageNet-22K pre-training and the much heavier architecture used in~\cite{jain2023oneformer}.
Since non-real-time state-of-the-art methods, such as OneFormer, do not report runtimes, we compare the number of floating point operations for input images of $1024\times 2048$ pixels.
Using this metric, our method outperforms all other methods by a large margin, underlining that current state-of-the-art methods primarily focus on accuracy rather than inference speed.
For example, OneFormer requires 497G floating point operations compared to 155G required for our method.

\begin{table}[!t]
\caption{\textbf{Comparison to non-real-time state-of-the-art panoptic segmentation methods on the Cityscapes dataset.} Methods marked with\,\dag\, use Mapillary Vistas~\cite{neuhold2017mapillary} pre-training, backbones marked with\,* use ImageNet-22K~\cite{deng2009imagenet} pre-training.}
\label{tab:cityscapes_sota_comp}
\captionsetup{labelfont=normalsize,textfont=normalsize,position=top}
\centering
\begin{tabular}{l|c|c|c|c|c}
    \toprule
    Method & Backbone & PQ $\uparrow$ & PQ$_{\text{th}}$ $\uparrow$ & PQ$_{\text{st}}$ $\uparrow$ & \#FLOPs $\downarrow$ \\
    \midrule
    RT-K-Net~\cite{schoen2023rt} & RTFormer~\cite{wang2022rtformer} & 60.2 & 51.5 & 66.5 & - \\
    \midrule
    Mask2Former~\cite{cheng2022masked} & ResNet-50~\cite{he2016deep} & 62.1 & - & - & - \\
    PanopticDepth~\cite{gao2022panopticdepth}\dag & ResNet-50~\cite{he2016deep} & 64.1 & 58.8 & 68.1 & - \\
    kMaX-DeepLab~\cite{yu2022k} & ResNet-50~\cite{he2016deep} & 64.3 & 57.7 & 69.1 & \underline{434G} \\
    Axial-DeepLab-XL~\cite{wang2020axial} & Axial-ResNet-XL~\cite{wang2020axial} & 64.4 & - & - & 2447G \\
    Mask2Former~\cite{cheng2022masked} & Swin-L~\cite{liu2021swin}* & 66.6 & - & - & 514G \\
    OneFormer~\cite{jain2023oneformer} & Swin-L~\cite{liu2021swin}* & 67.2 & - & - & 543G \\
    Axial-DeepLab-XL~\cite{wang2020axial}\dag & Axial-ResNet-XL~\cite{wang2020axial} & 67.8 & - & - & 2447G \\
    kMaX-DeepLab~\cite{yu2022k} & ConvNeXt-L~\cite{liu2022convnet}* & \underline{68.4} & \underline{62.9} & \underline{72.4} & 1673G \\
    OneFormer~\cite{jain2023oneformer}\dag & ConvNeXt-L~\cite{liu2022convnet}* & \textbf{70.1} & \textbf{64.6} & \textbf{74.1} & 497G \\
    \midrule
    \textbf{Ours} & RTFormer~\cite{wang2022rtformer} & 61.7 & 54.6 & 66.8 & \textbf{155G} \\
    \textbf{Ours}\dag & RTFormer~\cite{wang2022rtformer} & 64.0 & 56.3 & 69.5 & \textbf{155G} \\
    \bottomrule
\end{tabular}
\end{table}

\section{Ablation Study on Kernel Linking}
\label{sec:supp_abl_kernel_linking}
We perform an additional ablation study to investigate the effect of the \gls{kl} module.
As shown in \cref{tab:abl_kernel_linking}, \gls{kl} improves both the performance of panoptic segmentation and depth estimation.
This is in-line with results conducted in~\cite{yuan2022polyphonicformer, zhang2023unidaformer}, showing that a unified approach with explicit linking between both tasks can boost single-task performance.

\begin{table}[!t]
\caption{\textbf{Ablation study on \acrfull{kl} for MGNiceNet.} The ablation study is conducted on the Cityscapes~\cite{cordts2016cityscapes} validation dataset.}
\label{tab:abl_kernel_linking}
\captionsetup{labelfont=normalsize,textfont=normalsize,position=top}
\centering
\begin{tabular}{l|c|c|c} 
    \toprule
    Method & \acrfull{kl} & PQ $\uparrow$  & RMSE $\downarrow$ \\
    \midrule
    \multirow{2}{*}{Ours} & -  & 63.4 & 7.3 \\ 
    & \checkmark & 64.0 & 7.1 \\
    \bottomrule
\end{tabular}
\end{table}

\section{Qualitative Results}
\label{sec:supp_qualitative}
\begin{figure}[t]
\begin{subfigure}[b]{0.24\textwidth}
  \centering
  \includegraphics[width=\textwidth]{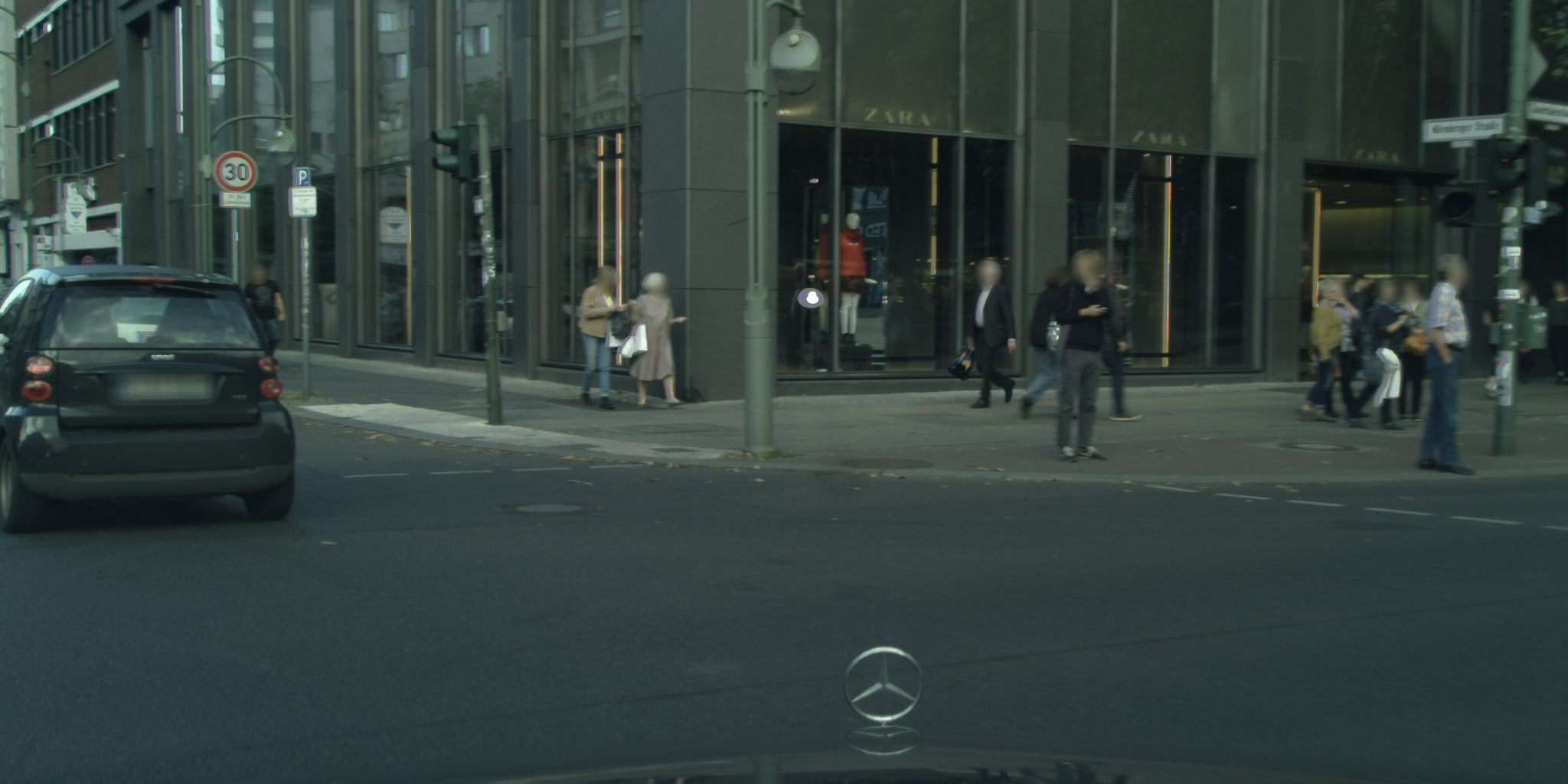}
\end{subfigure}
\hfill
\begin{subfigure}[b]{0.24\textwidth}
  \centering
  \includegraphics[width=\textwidth]{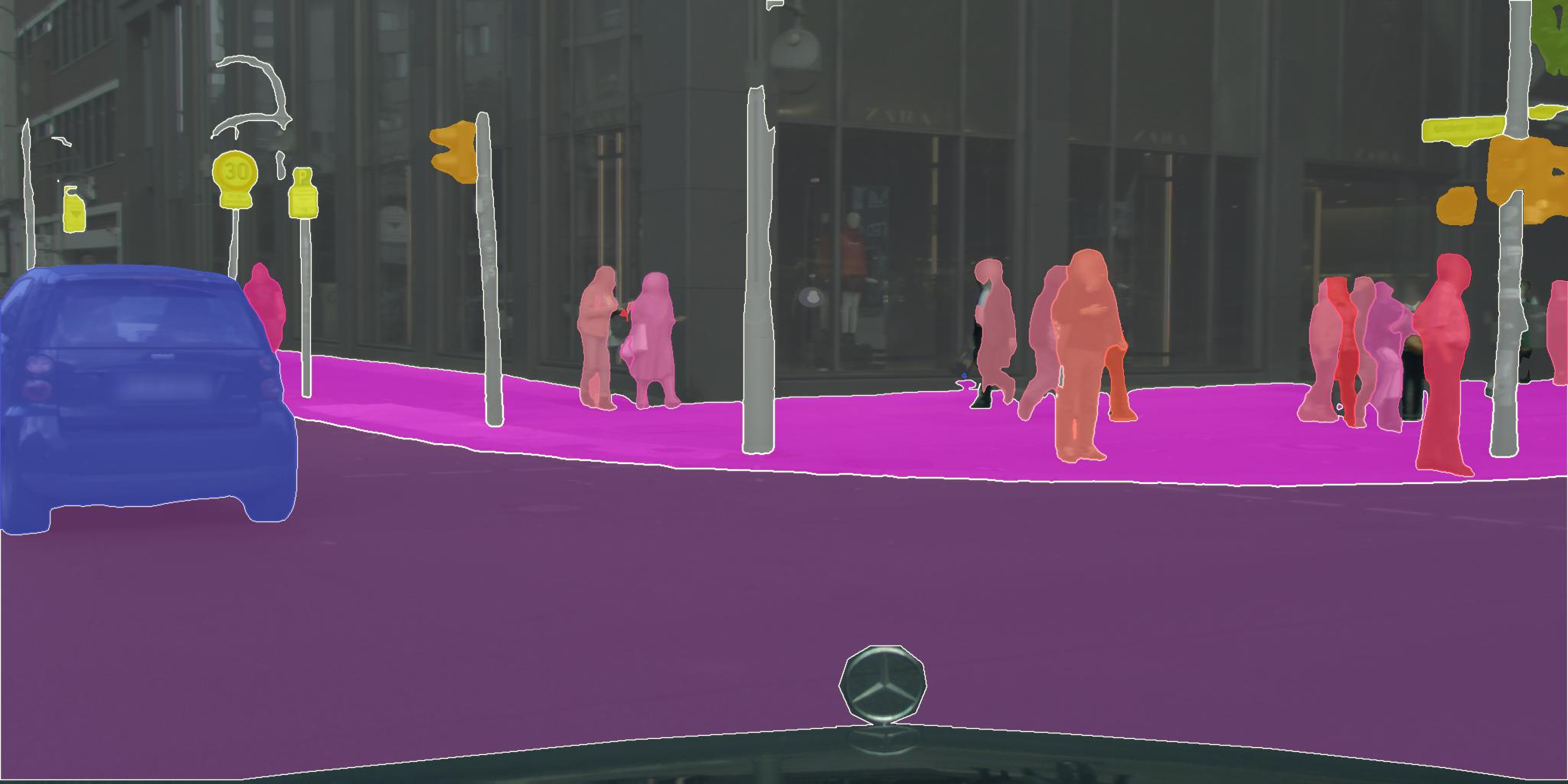}
\end{subfigure}
\hfill
\begin{subfigure}[b]{0.24\textwidth}
  \centering
  \includegraphics[width=\textwidth]{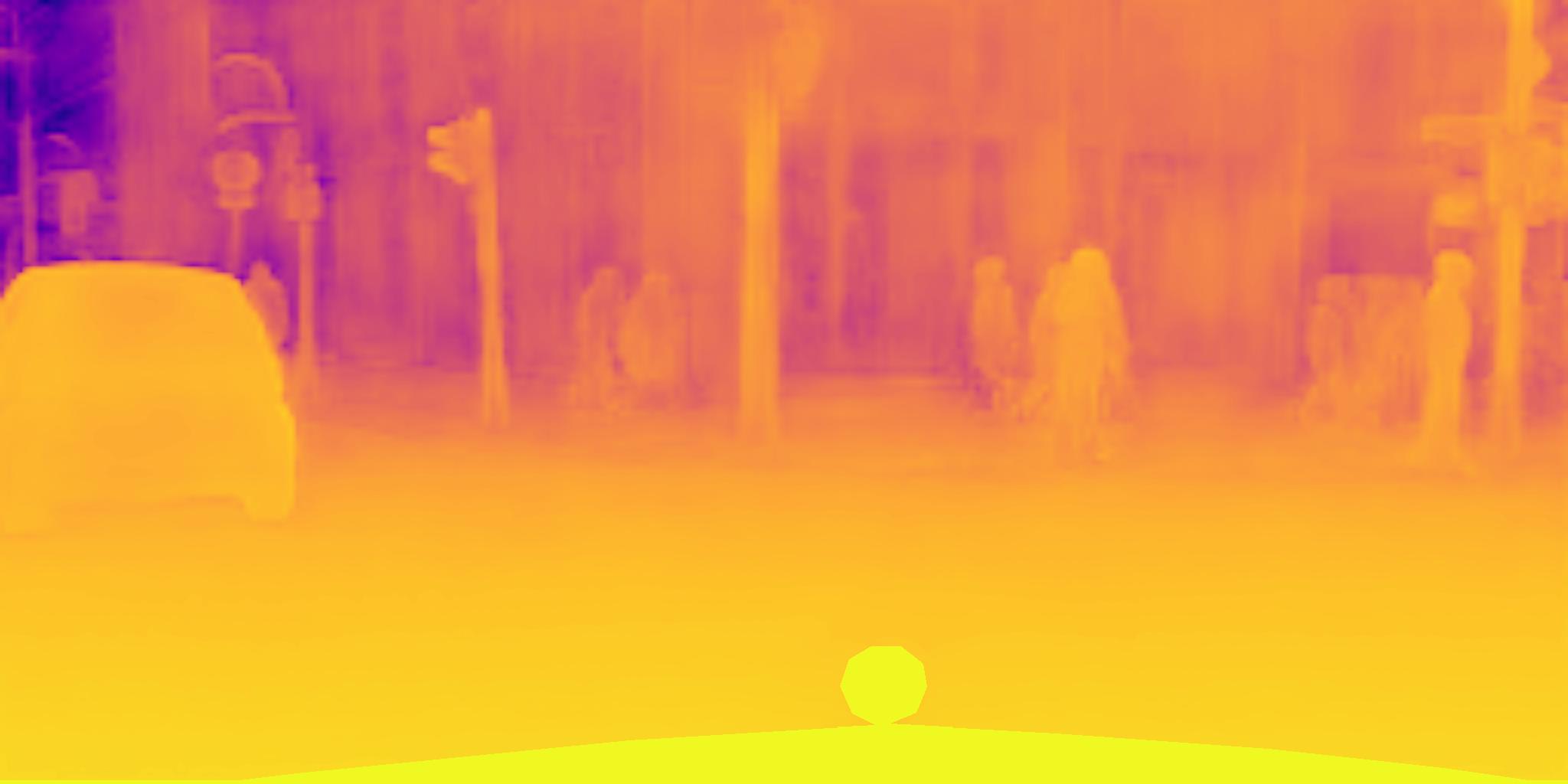}
\end{subfigure}
\hfill
\begin{subfigure}[b]{0.24\textwidth}
  \centering
  \includegraphics[width=\textwidth]{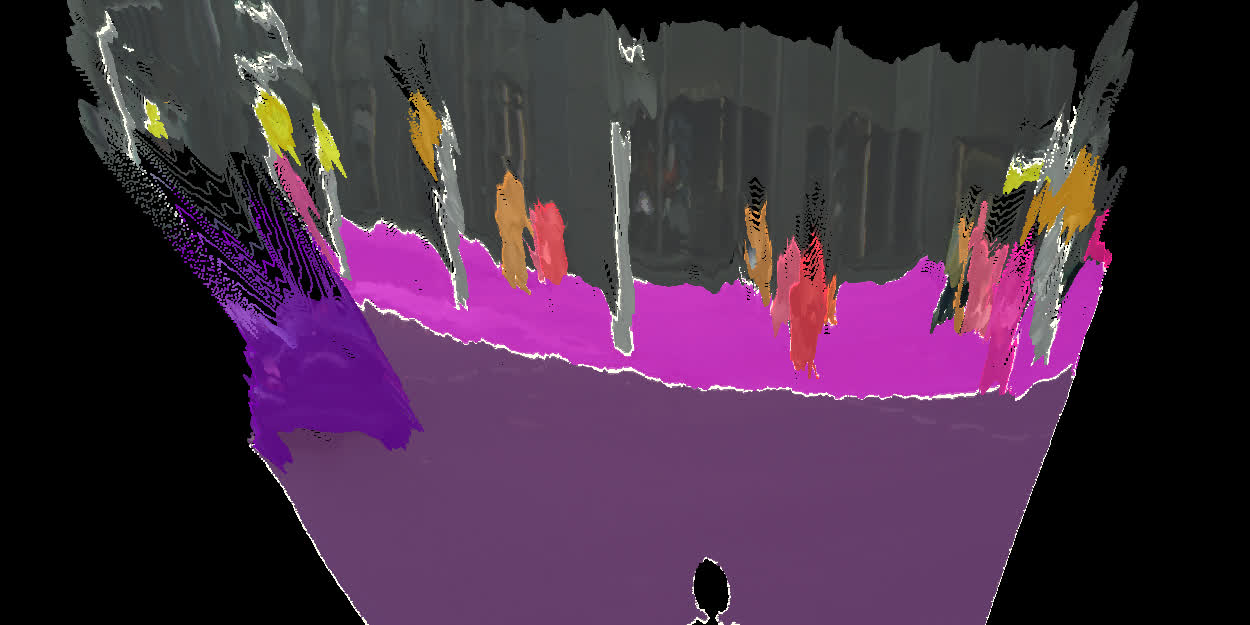}
\end{subfigure}
\hfill
\begin{subfigure}[b]{0.24\textwidth}
  \centering
  \includegraphics[width=\textwidth]{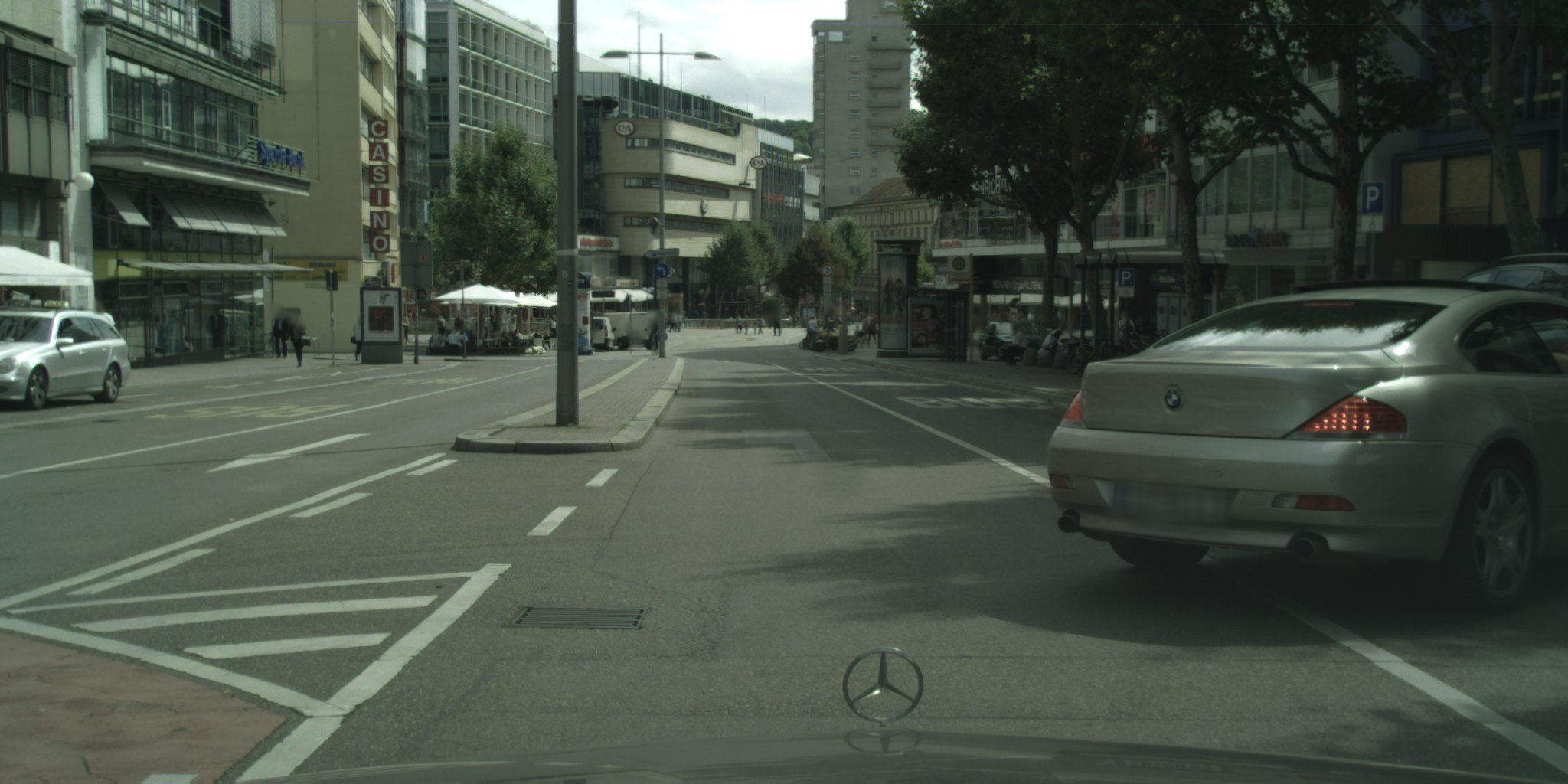}
\end{subfigure}
\hfill
\begin{subfigure}[b]{0.24\textwidth}
  \centering
  \includegraphics[width=\textwidth]{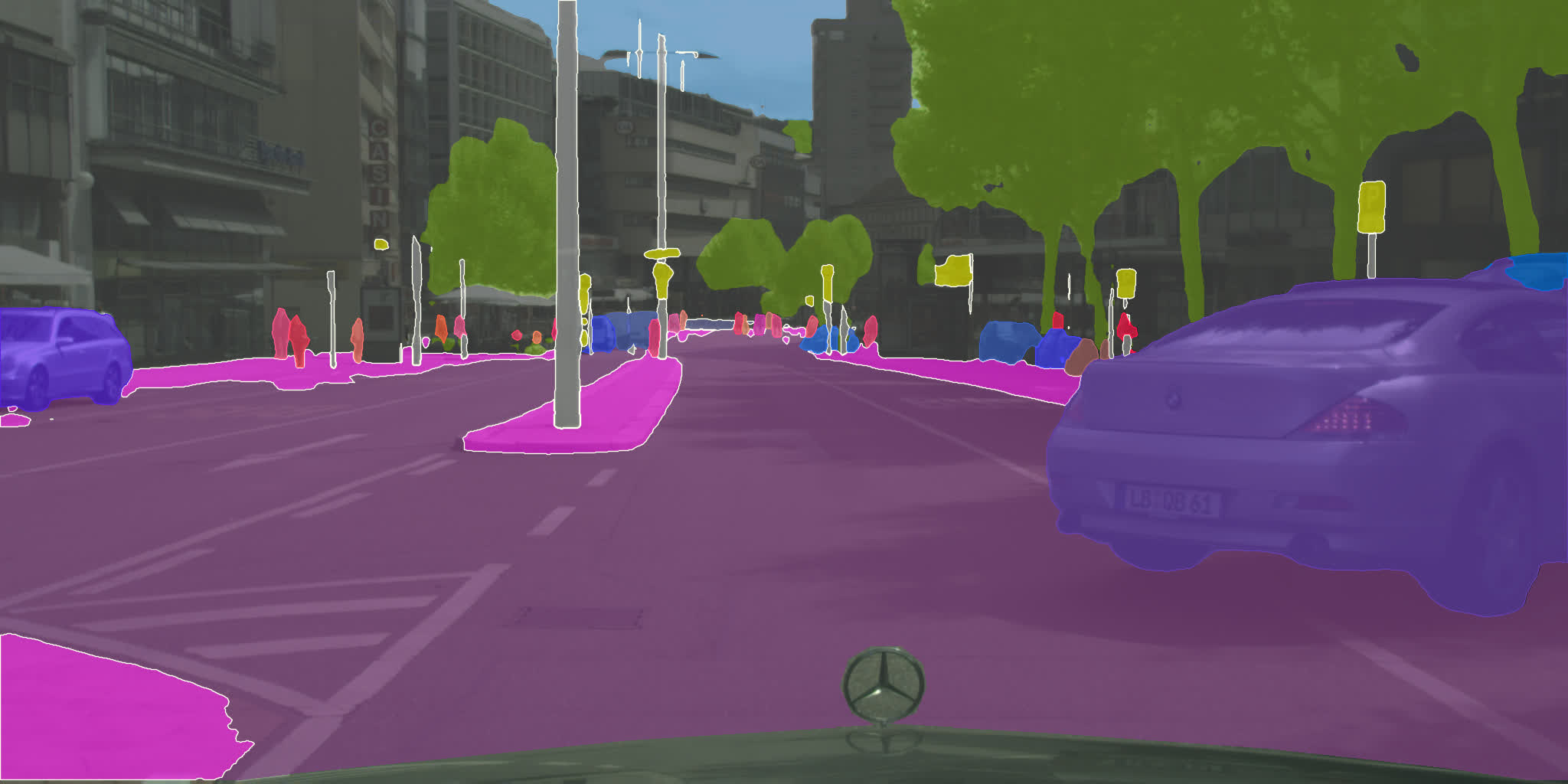}
\end{subfigure}
\hfill
\begin{subfigure}[b]{0.24\textwidth}
  \centering
  \includegraphics[width=\textwidth]{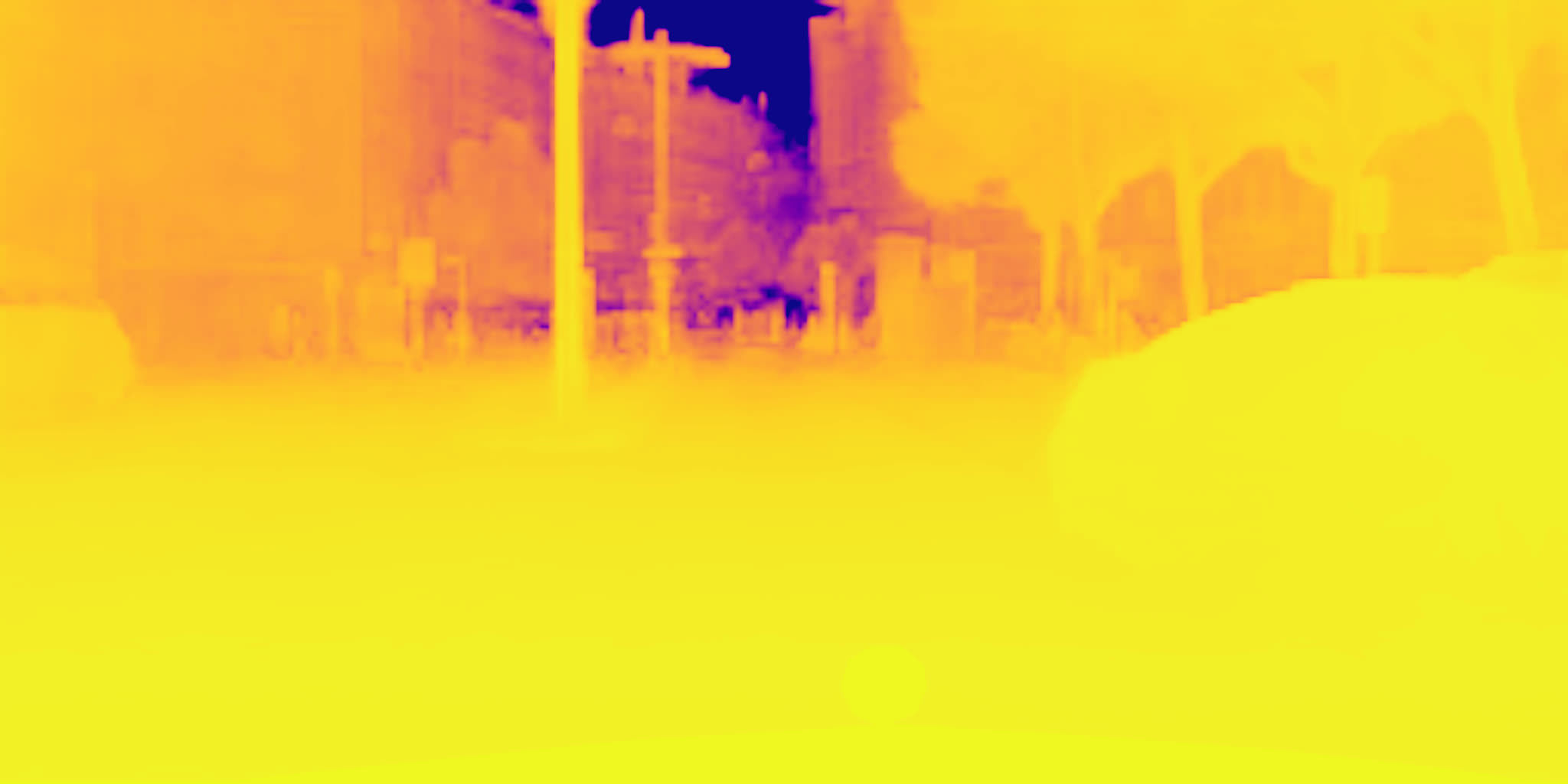}
\end{subfigure}
\hfill
\begin{subfigure}[b]{0.24\textwidth}
  \centering
  \includegraphics[width=\textwidth]{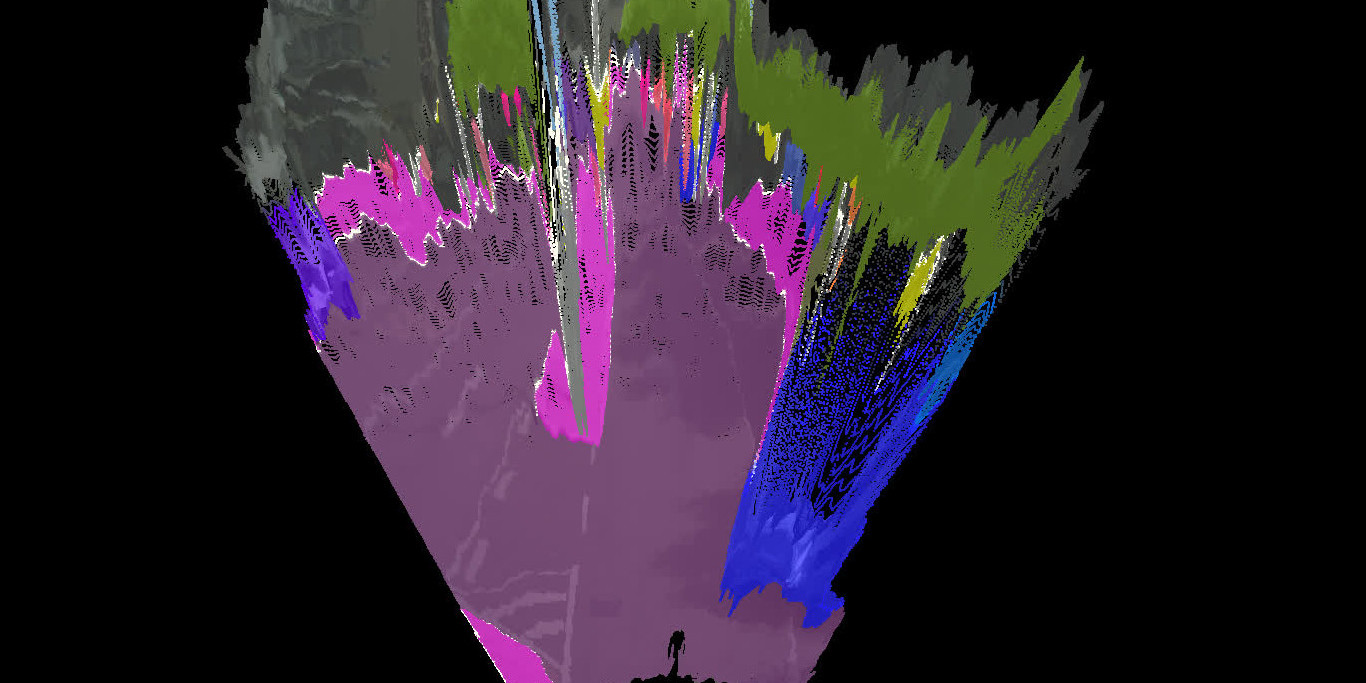}
\end{subfigure}
\hfill
\begin{subfigure}[b]{0.24\textwidth}
  \centering
  \includegraphics[width=\textwidth]{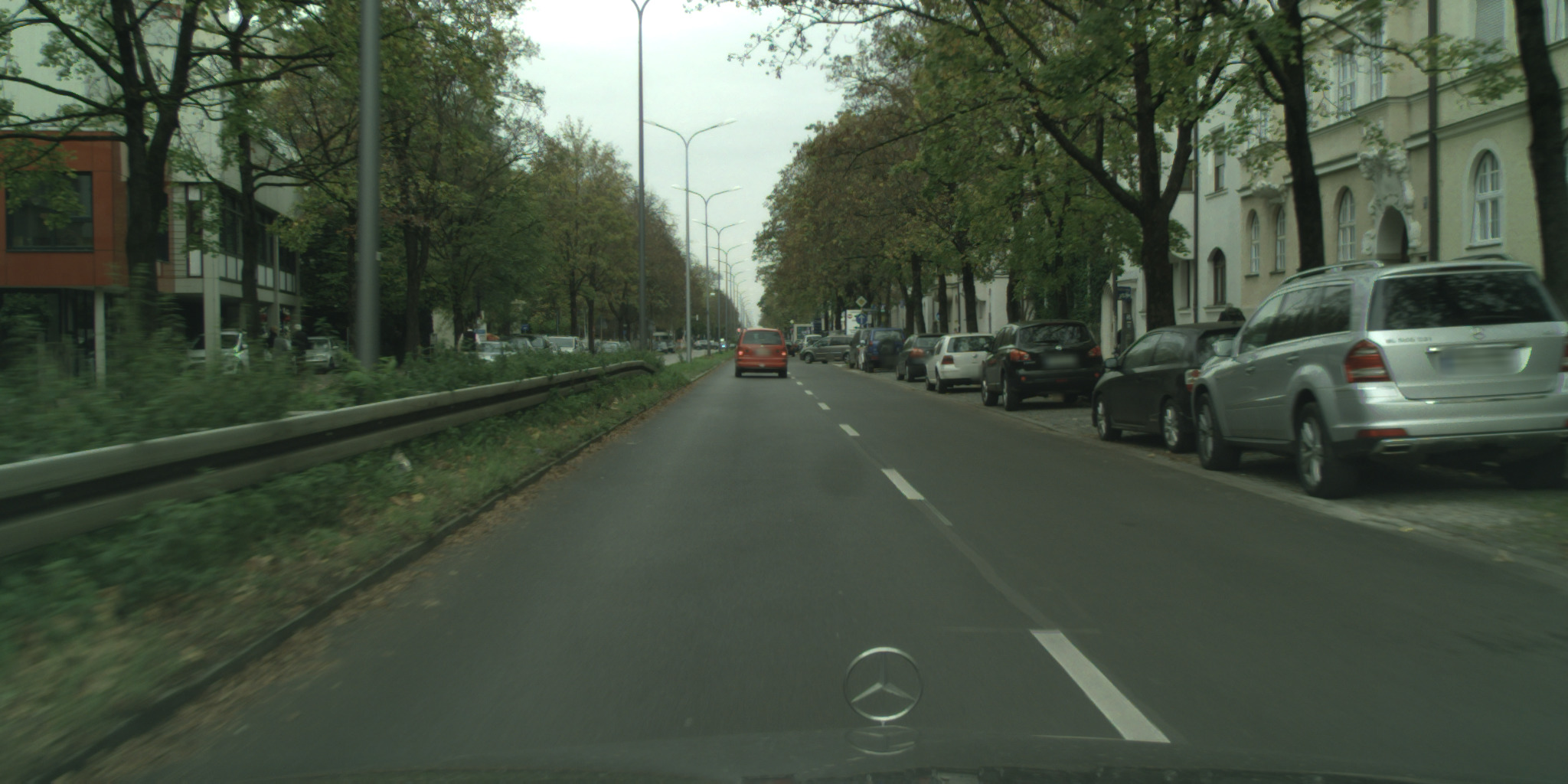}
\end{subfigure}
\hfill
\begin{subfigure}[b]{0.24\textwidth}
  \centering
  \includegraphics[width=\textwidth]{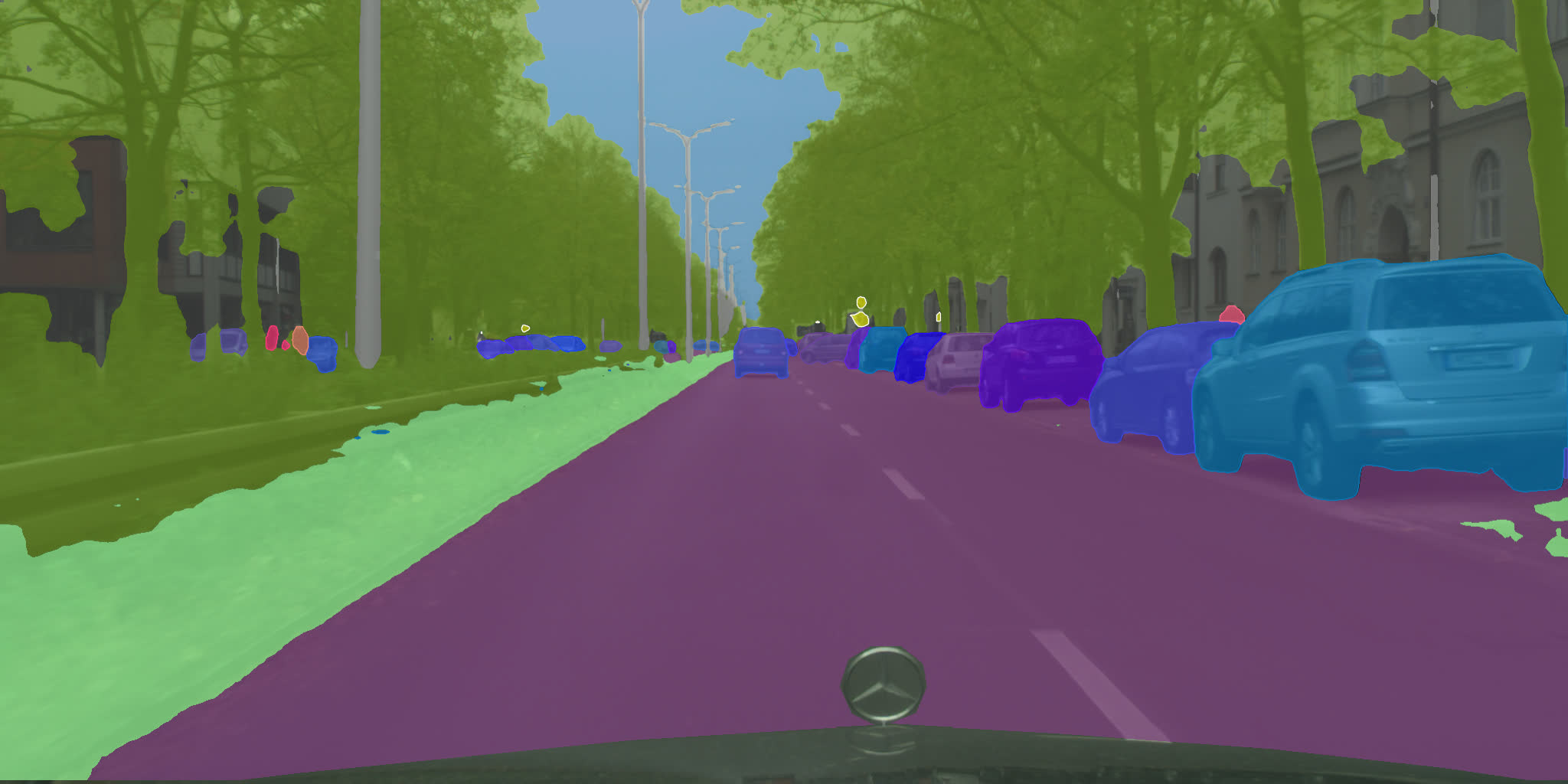}
\end{subfigure}
\hfill
\begin{subfigure}[b]{0.24\textwidth}
  \centering
  \includegraphics[width=\textwidth]{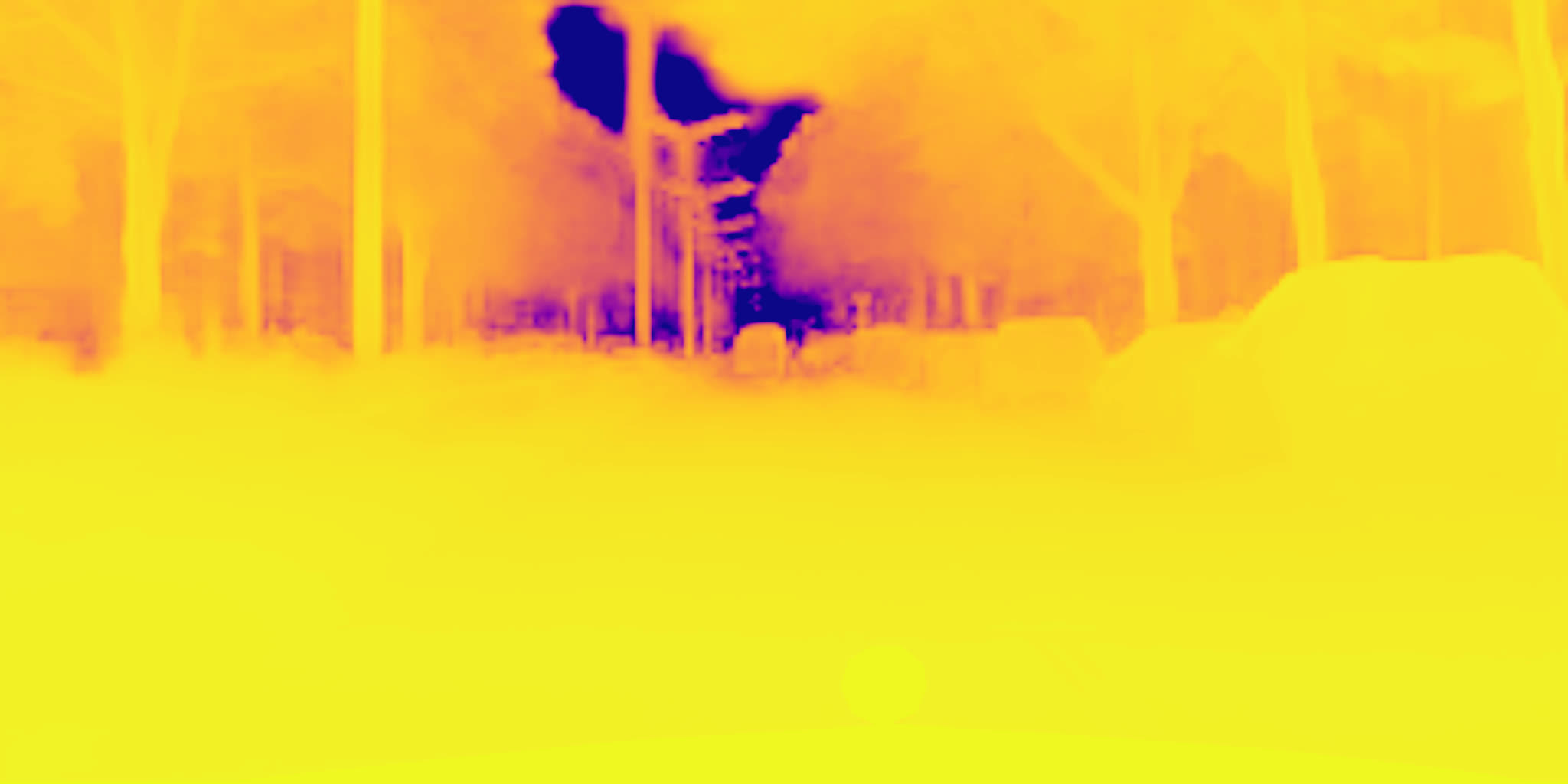}
\end{subfigure}
\hfill
\begin{subfigure}[b]{0.24\textwidth}
  \centering
  \includegraphics[width=\textwidth]{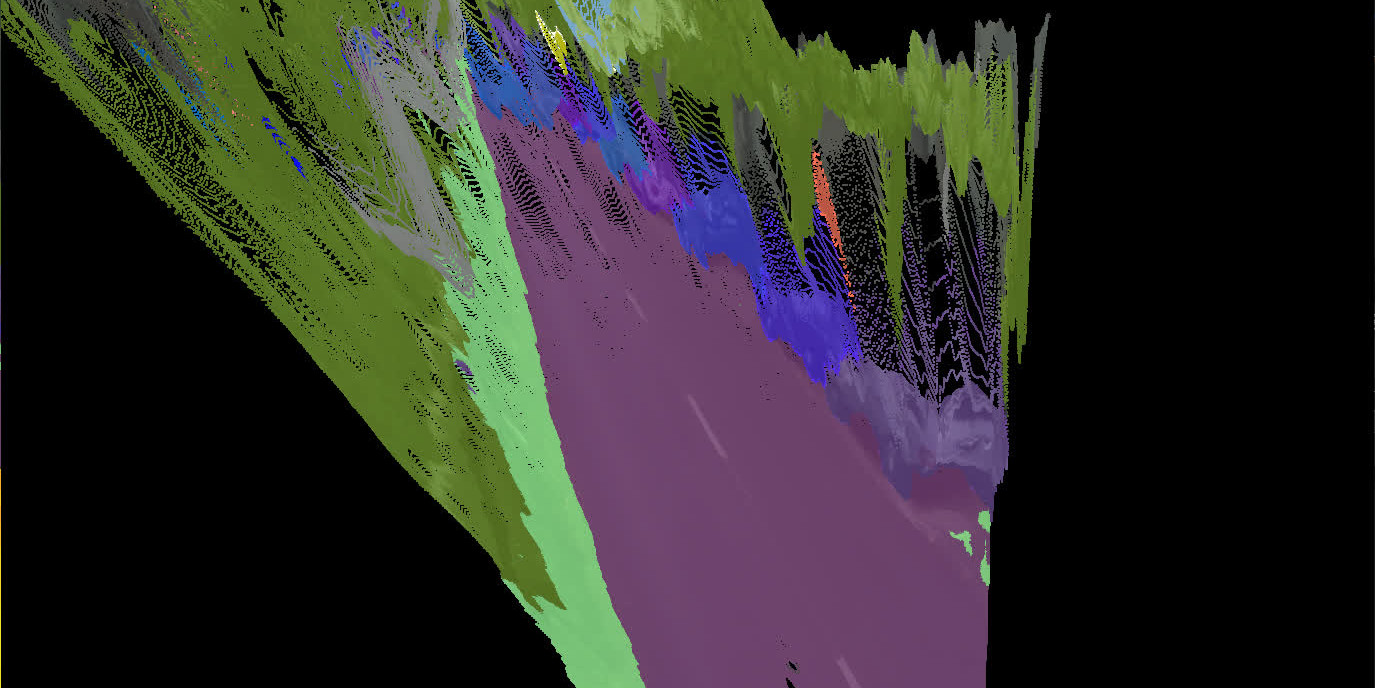}
\end{subfigure}
\vspace{0.2cm}
\begin{subfigure}[b]{0.24\textwidth}
  \centering
  \includegraphics[width=\textwidth]{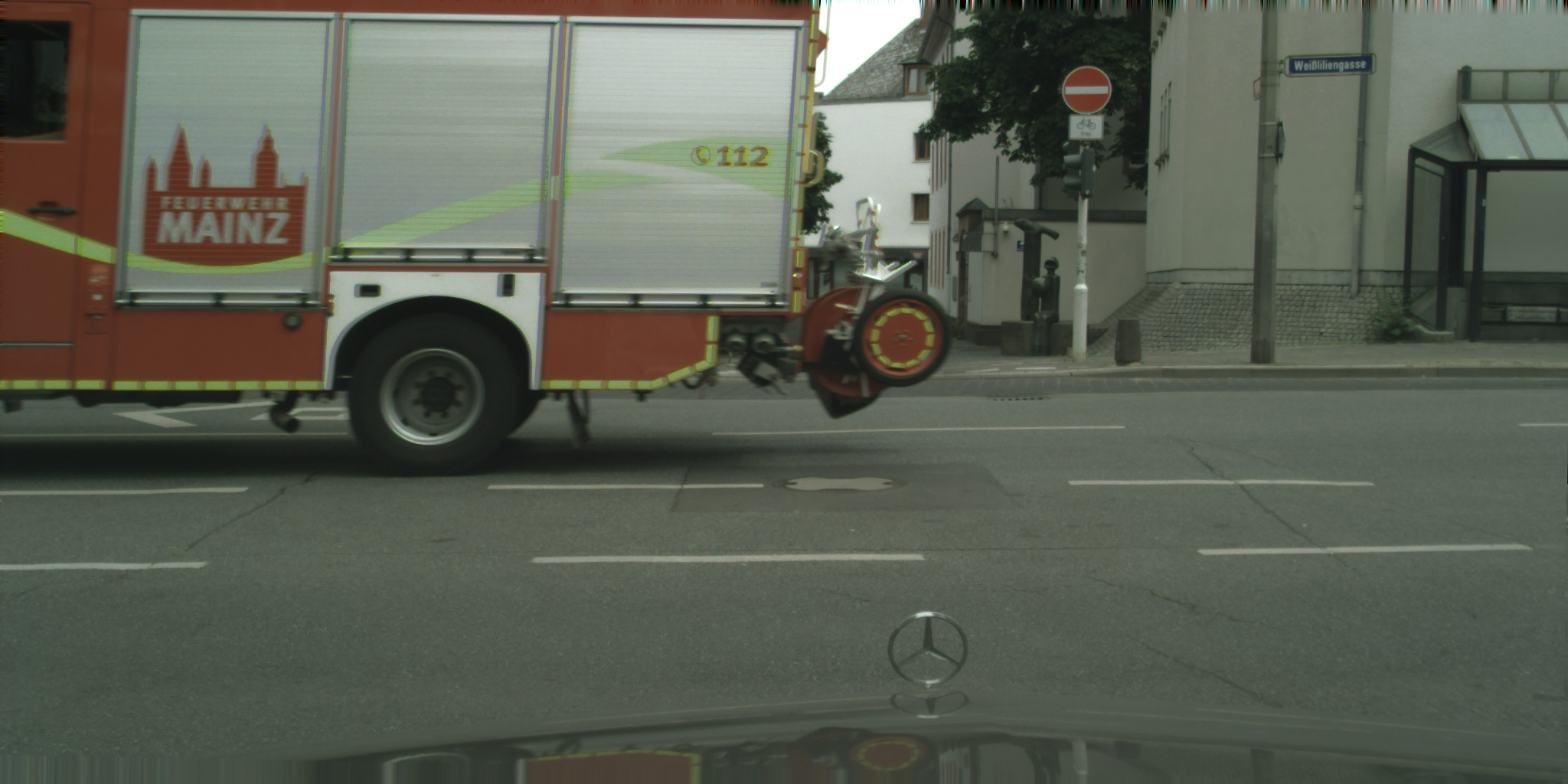}
\end{subfigure}
\hfill
\begin{subfigure}[b]{0.24\textwidth}
  \centering
  \includegraphics[width=\textwidth]{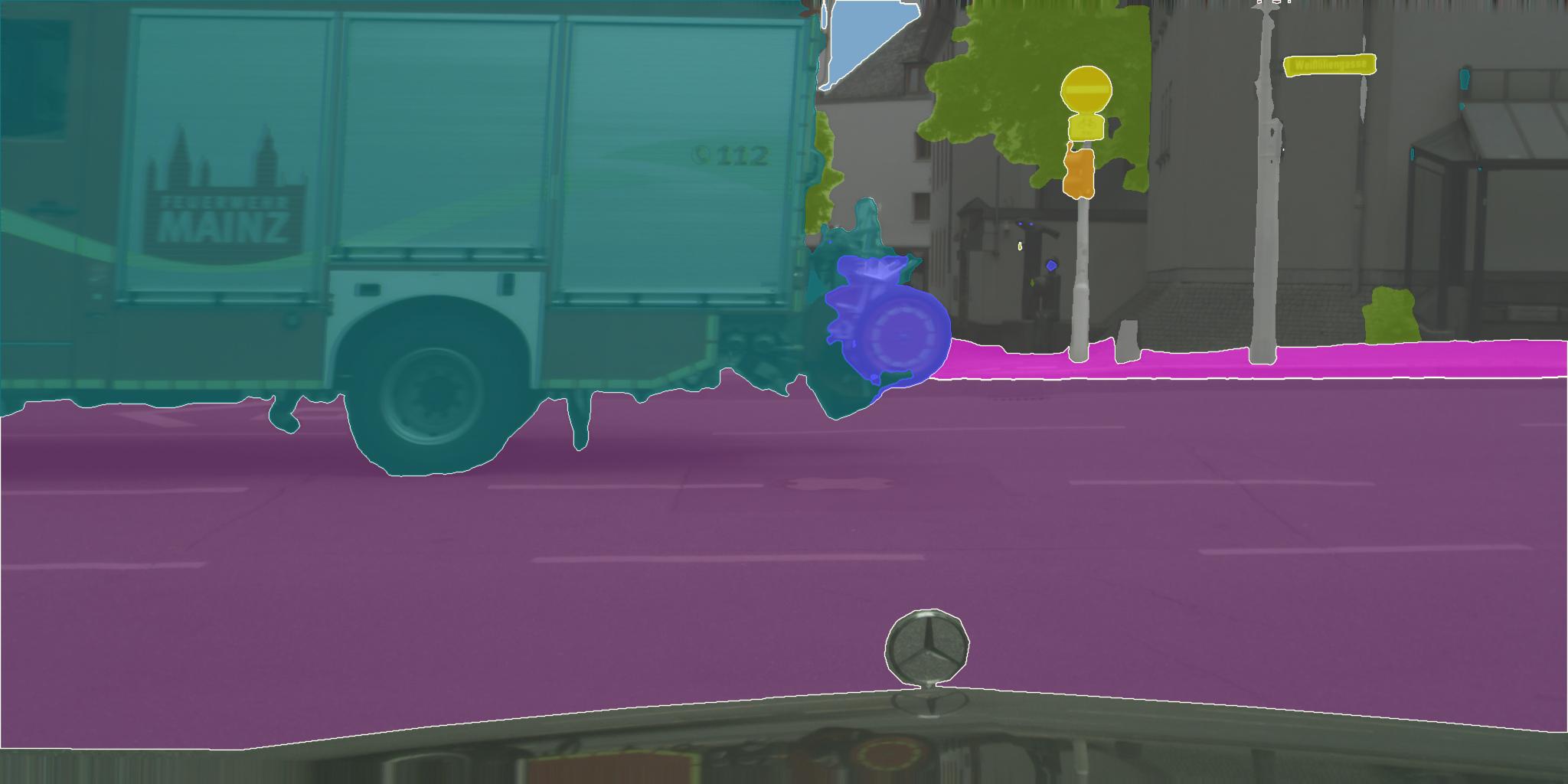}
\end{subfigure}
\hfill
\begin{subfigure}[b]{0.24\textwidth}
  \centering
  \includegraphics[width=\textwidth]{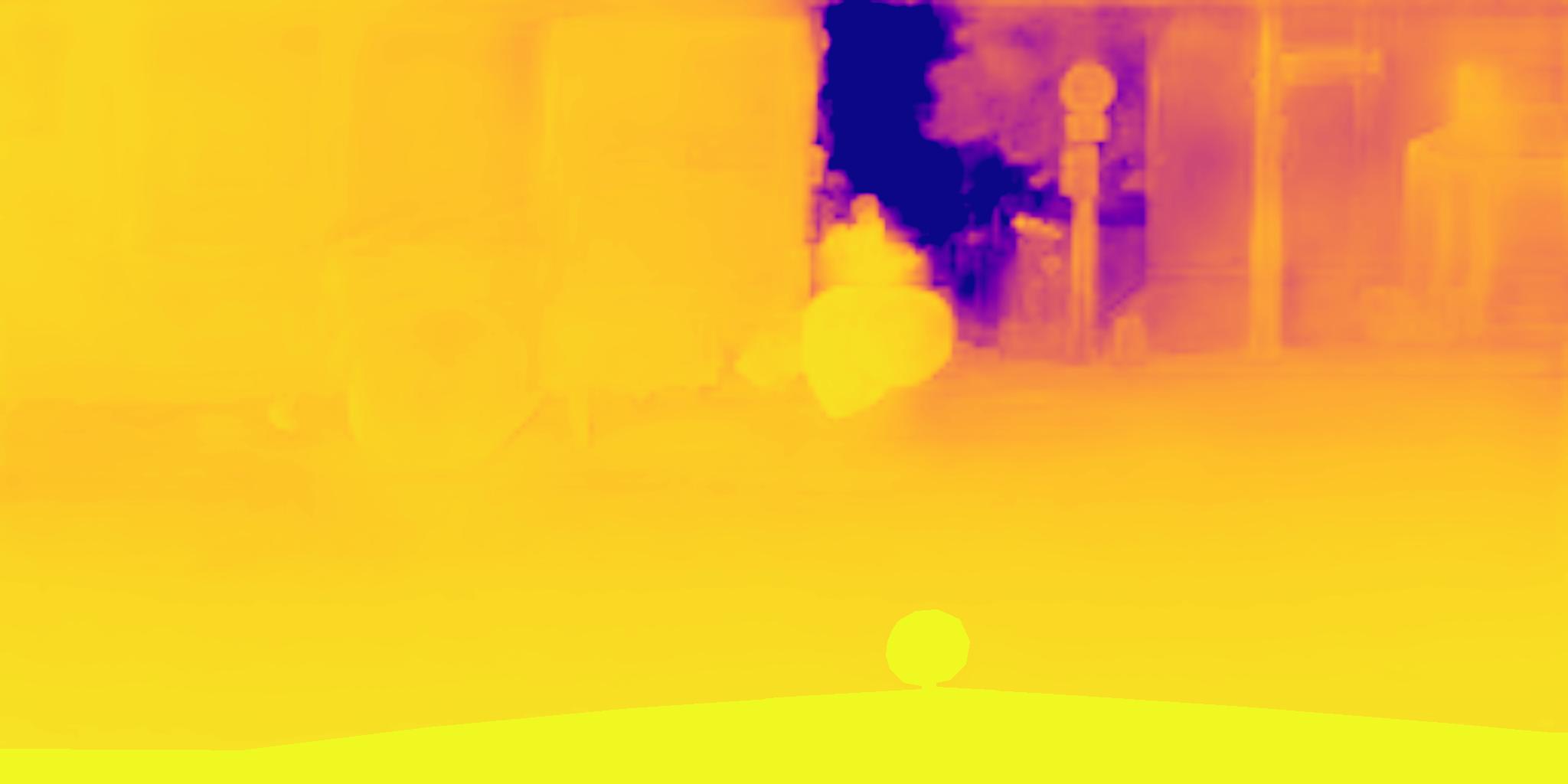}
\end{subfigure}
\hfill
\begin{subfigure}[b]{0.24\textwidth}
  \centering
  \includegraphics[width=\textwidth]{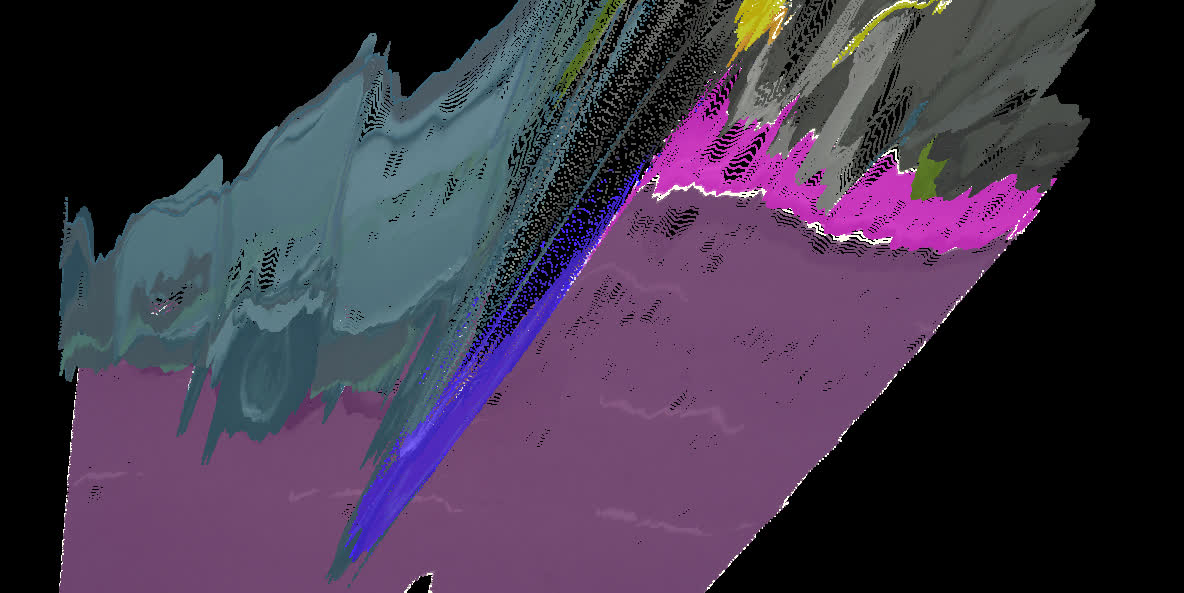}
\end{subfigure}
\hfill
\begin{subfigure}[b]{0.24\textwidth}
  \centering
  \includegraphics[width=\textwidth]{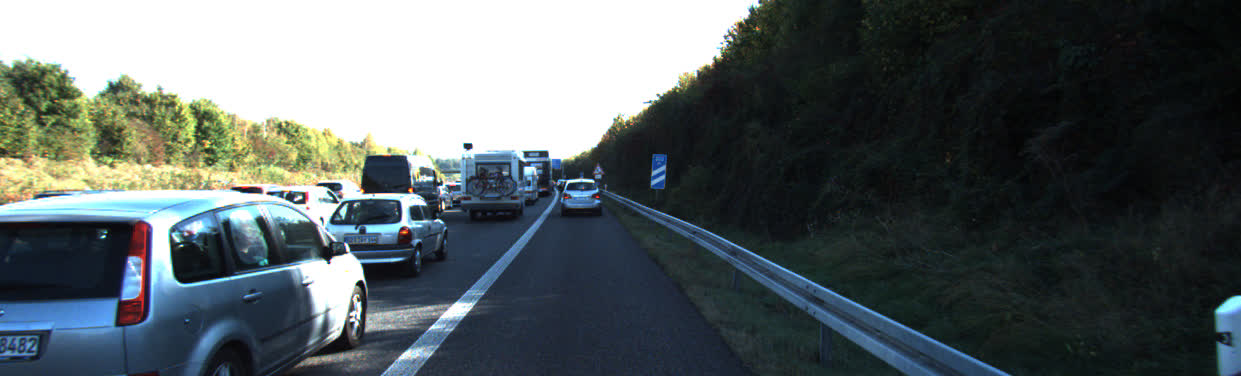}
\end{subfigure}
\hfill
\begin{subfigure}[b]{0.24\textwidth}
  \centering
  \includegraphics[width=\textwidth]{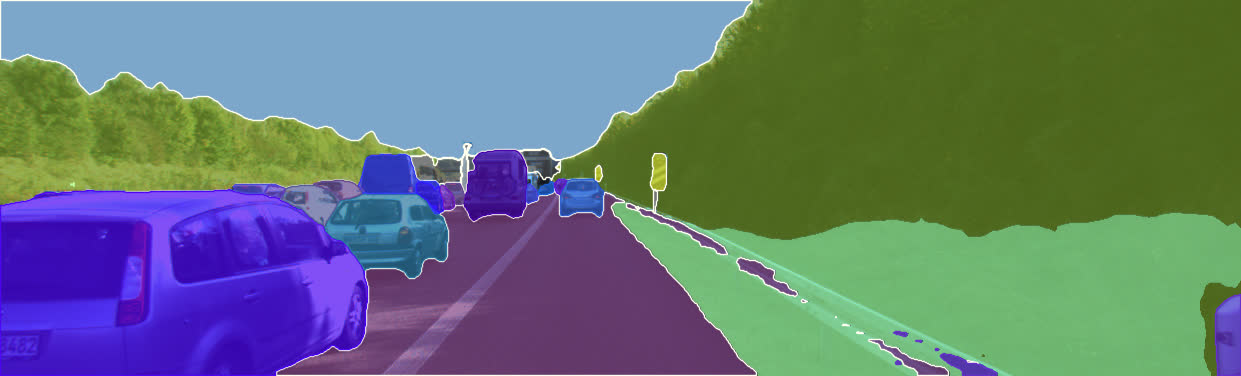}
\end{subfigure}
\hfill
\begin{subfigure}[b]{0.24\textwidth}
  \centering
  \includegraphics[width=\textwidth]{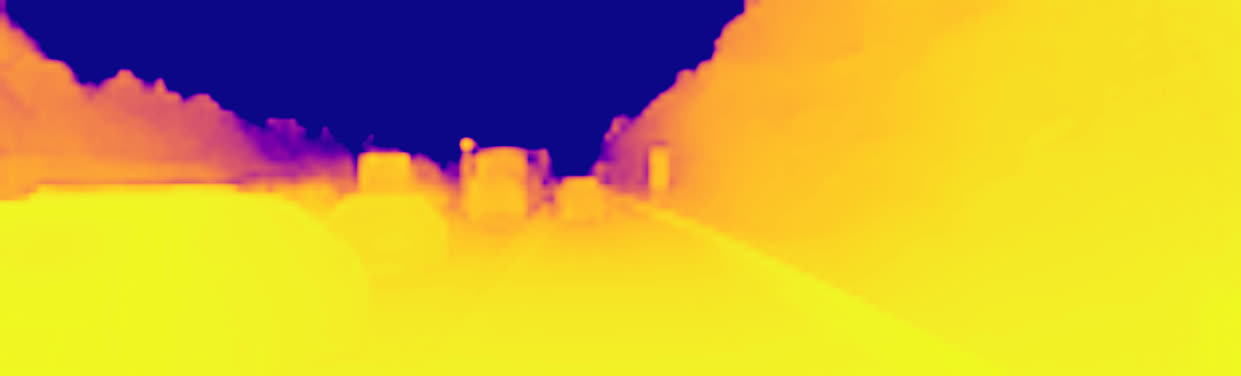}
\end{subfigure}
\hfill
\begin{subfigure}[b]{0.24\textwidth}
  \centering
  \includegraphics[width=\textwidth]{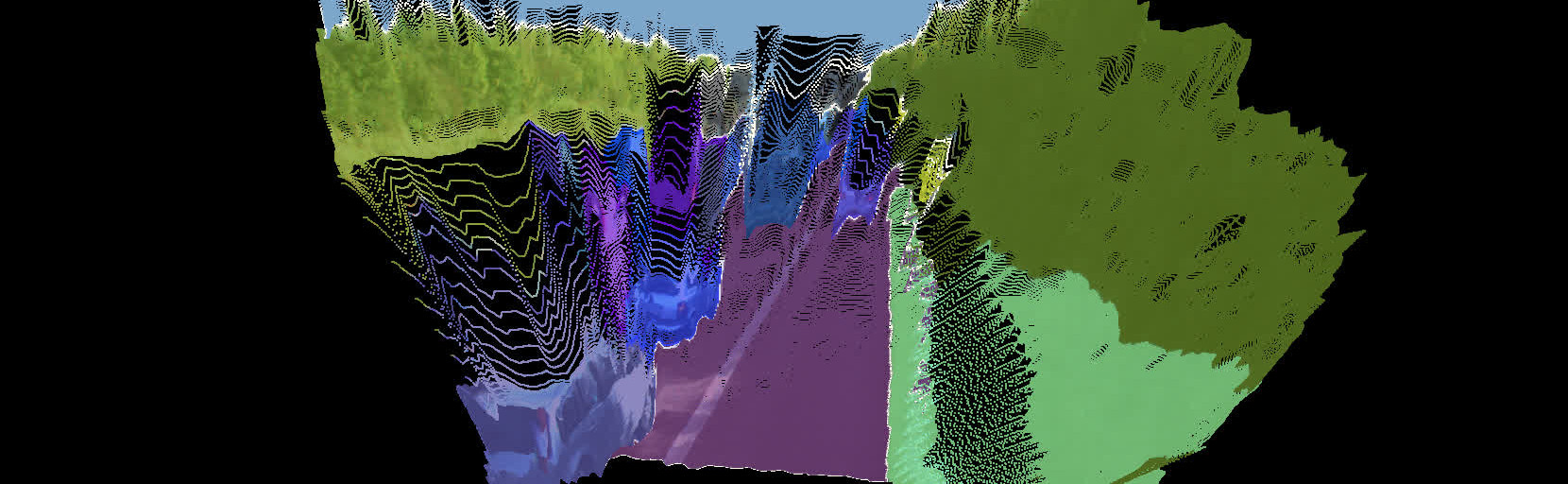}
\end{subfigure}
\hfill
\begin{subfigure}[b]{0.24\textwidth}
  \centering
  \includegraphics[width=\textwidth]{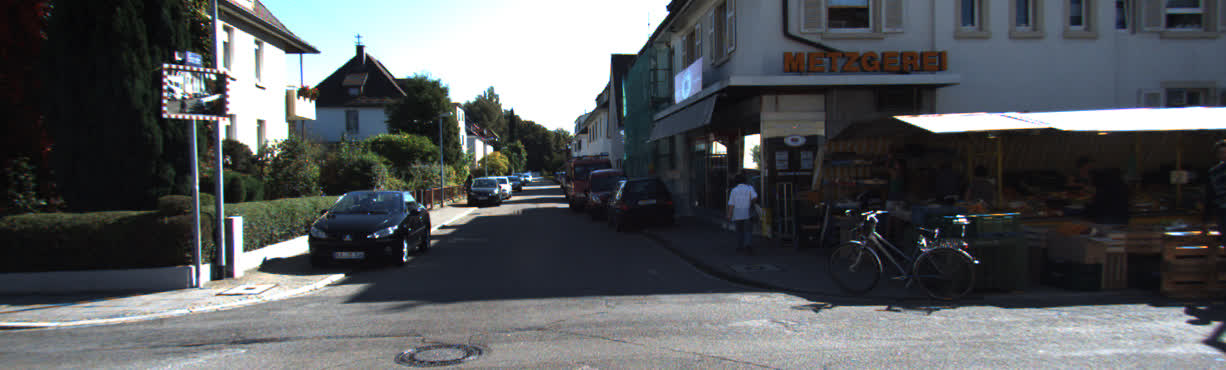}
\end{subfigure}
\hfill
\begin{subfigure}[b]{0.24\textwidth}
  \centering
  \includegraphics[width=\textwidth]{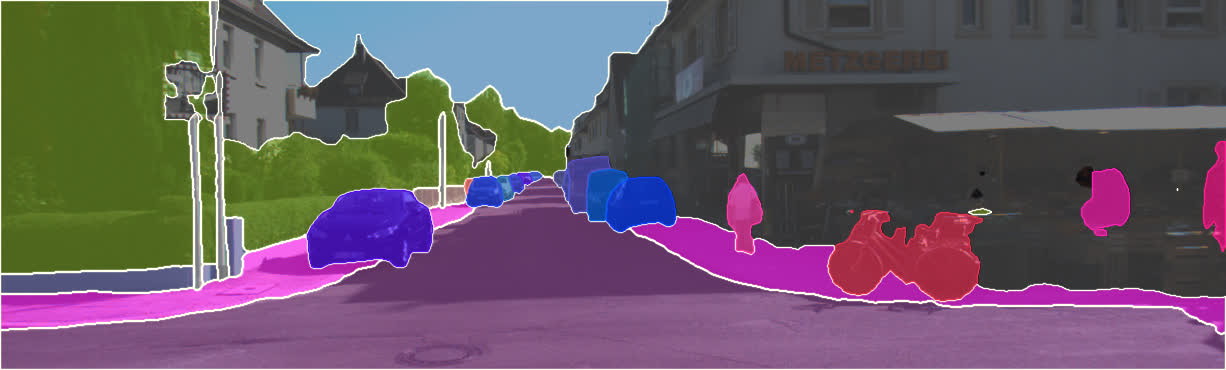}
\end{subfigure}
\hfill
\begin{subfigure}[b]{0.24\textwidth}
  \centering
  \includegraphics[width=\textwidth]{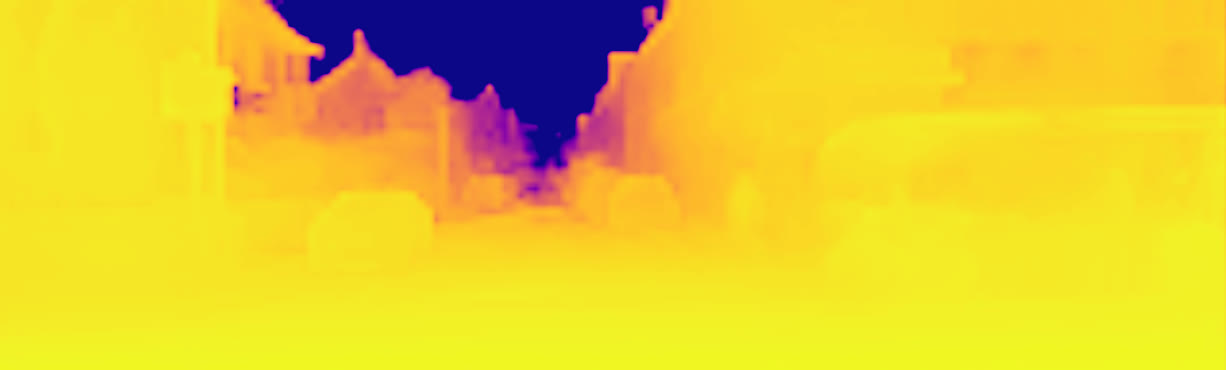}
\end{subfigure}
\hfill
\begin{subfigure}[b]{0.24\textwidth}
  \centering
  \includegraphics[width=\textwidth]{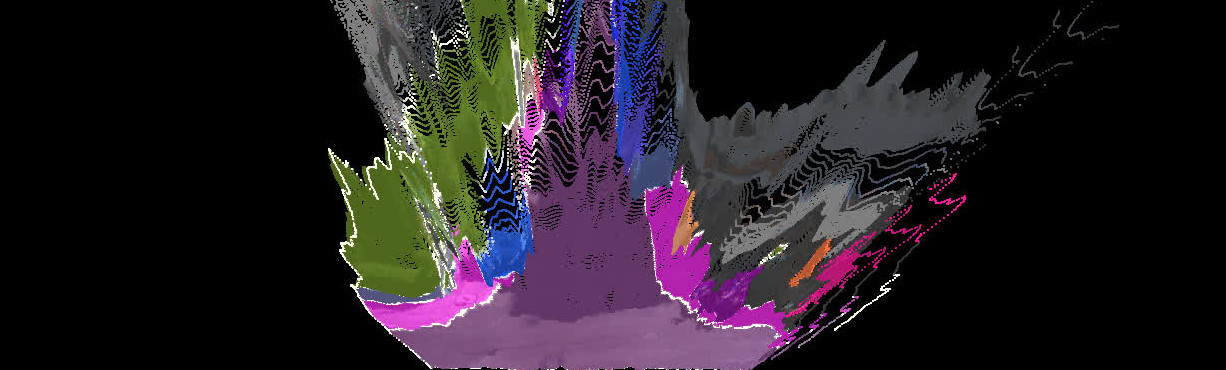}
\end{subfigure}
\hfill
\begin{subfigure}[b]{0.24\textwidth}
  \centering
  \includegraphics[width=\textwidth]{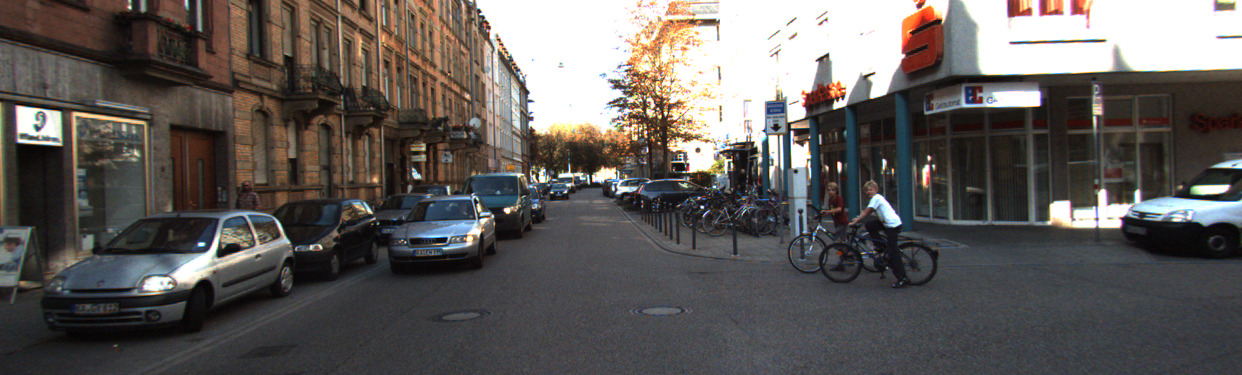}
\end{subfigure}
\hfill
\begin{subfigure}[b]{0.24\textwidth}
  \centering
  \includegraphics[width=\textwidth]{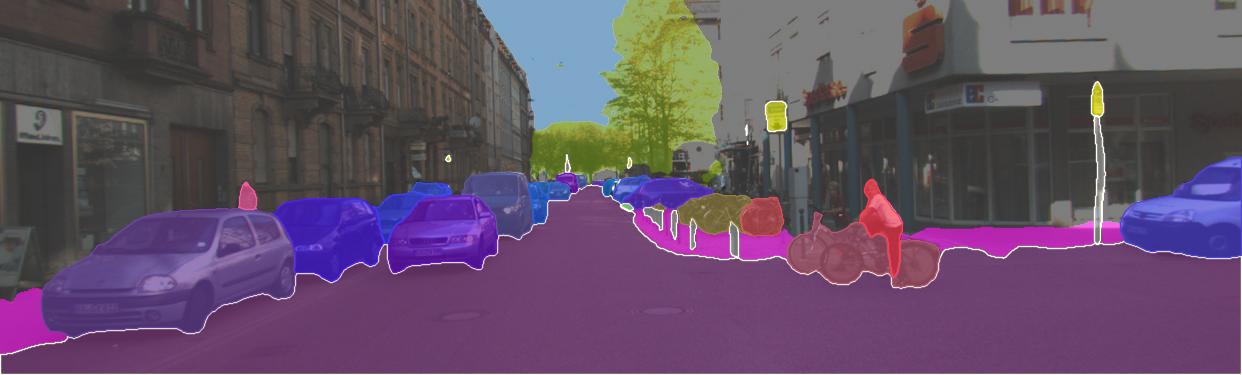}
\end{subfigure}
\hfill
\begin{subfigure}[b]{0.24\textwidth}
  \centering
  \includegraphics[width=\textwidth]{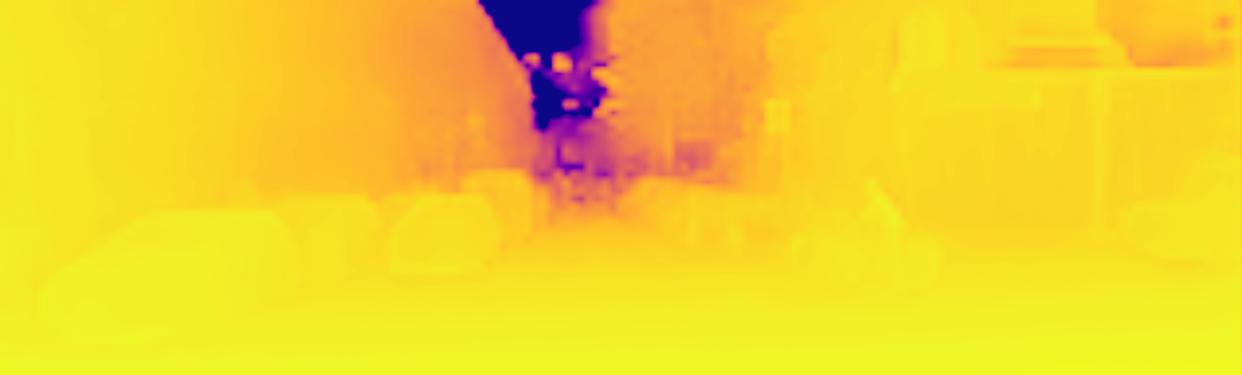}
\end{subfigure}
\hfill
\begin{subfigure}[b]{0.24\textwidth}
  \centering
  \includegraphics[width=\textwidth]{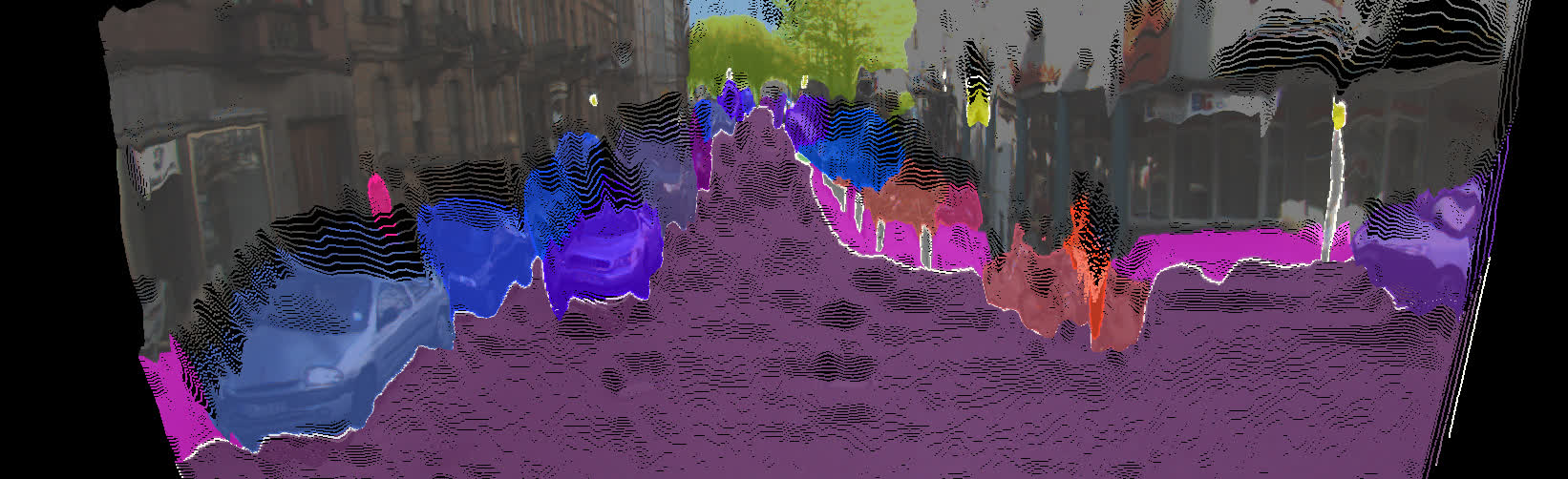}
\end{subfigure}
\hfill
\begin{subfigure}[b]{0.24\textwidth}
  \centering
  \includegraphics[width=\textwidth]{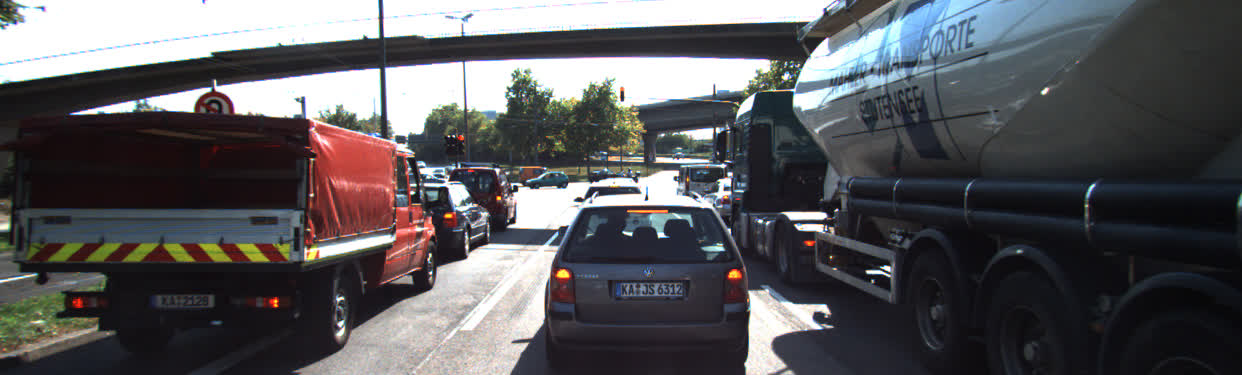}
\end{subfigure}
\hfill
\begin{subfigure}[b]{0.24\textwidth}
  \centering
  \includegraphics[width=\textwidth]{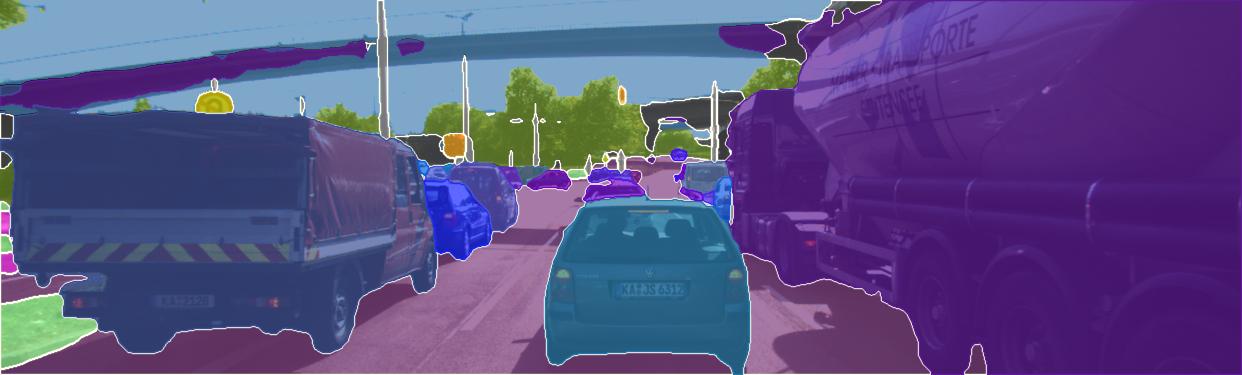}
\end{subfigure}
\hfill
\begin{subfigure}[b]{0.24\textwidth}
  \centering
  \includegraphics[width=\textwidth]{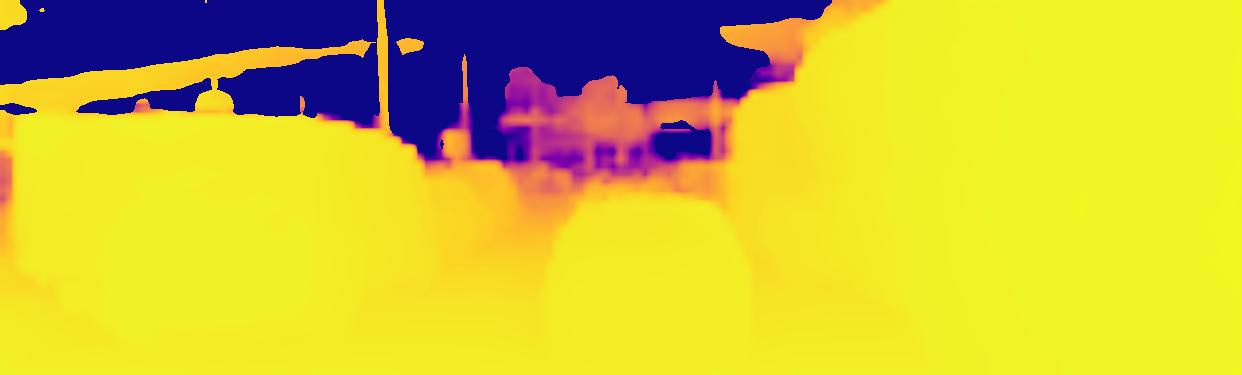}
\end{subfigure}
\hfill
\begin{subfigure}[b]{0.24\textwidth}
  \centering
  \includegraphics[width=\textwidth]{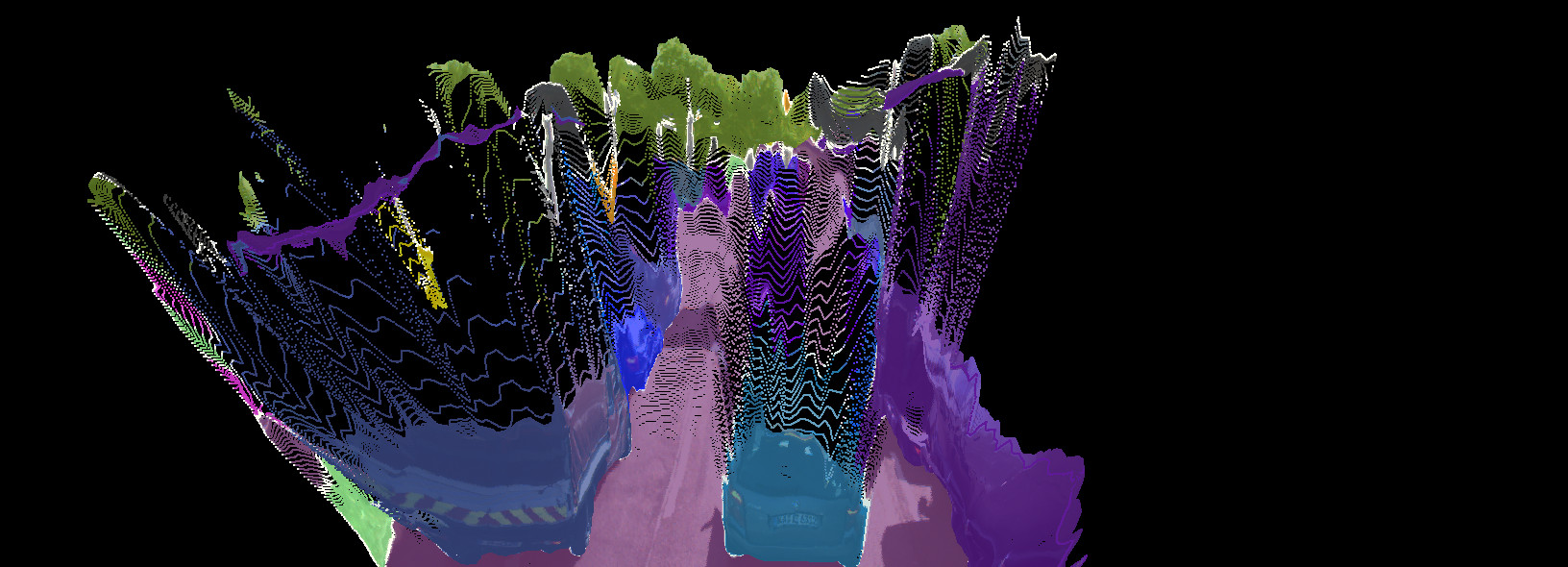}
\end{subfigure}
\caption{\textbf{Qualitative results of MGNiceNet on Cityscapes and KITTI.} The columns (from left to right) show the input image, the panoptic prediction overlay, the monocular depth estimation, and the 3D panoptic point cloud prediction, respectively. The top block shows predictions on the Cityscapes~\cite{cordts2016cityscapes} dataset, while the bottom block shows predictions on the KITTI~\cite{geiger2013vision} dataset. In each block, the last row shows an example of inaccurate predictions.}
\label{fig:qualitative}
\end{figure}

\begin{figure}[t]
\begin{subfigure}[b]{0.35\textwidth}
  \centering
  \begin{tikzpicture}
    \node[anchor=south west,inner sep=0] at (0,0) {\includegraphics[width=\textwidth]{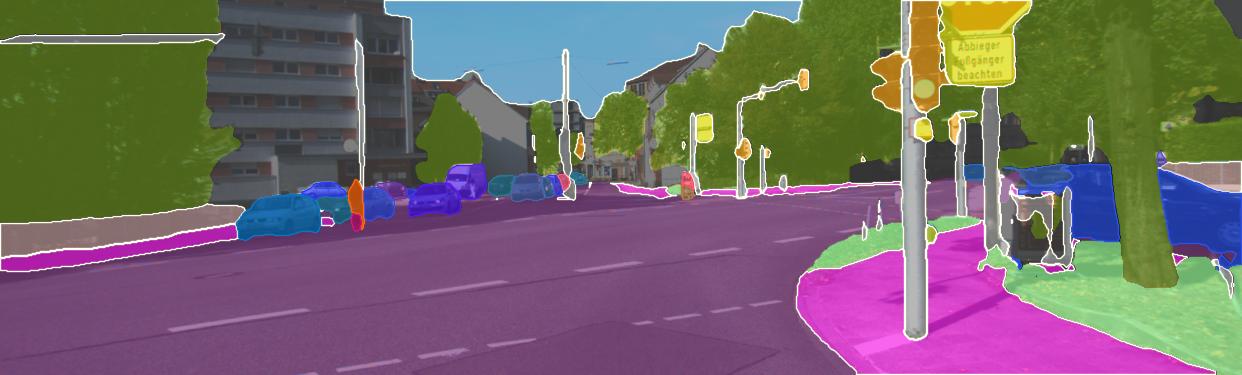}};
    \draw[red,thick] (1.15,0.45) rectangle (1.3,0.7);
    \draw[red,thick] (2.5,0.02) rectangle (4.26,0.35);
  \end{tikzpicture}
\end{subfigure}
\hfill
\begin{subfigure}[b]{0.35\textwidth}
  \centering
  \begin{tikzpicture}
    \node[anchor=south west,inner sep=0] at (0,0) {\includegraphics[width=\textwidth]{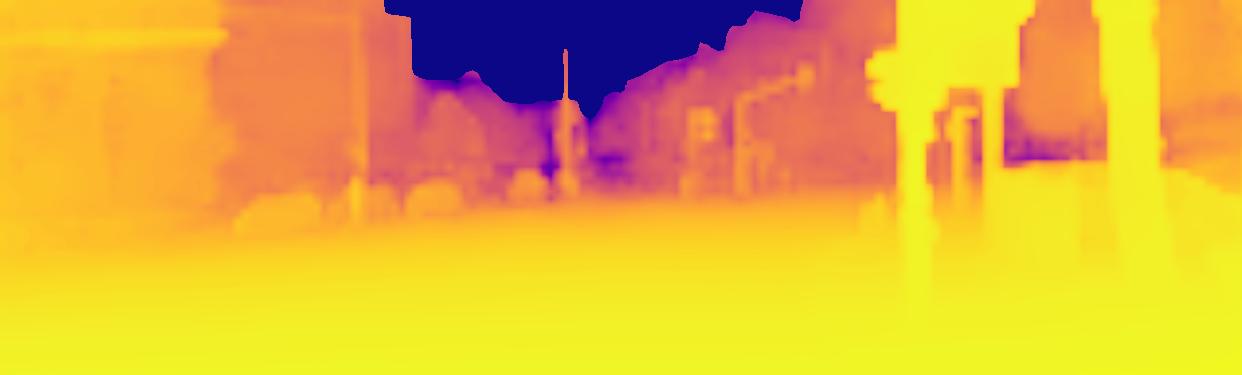}};
    \draw[red,thick] (2.3,0.55) rectangle (2.9,1.1);
  \end{tikzpicture}
\end{subfigure}
\hfill
\begin{subfigure}[b]{0.257\textwidth}
  \centering
  \includegraphics[width=\textwidth]{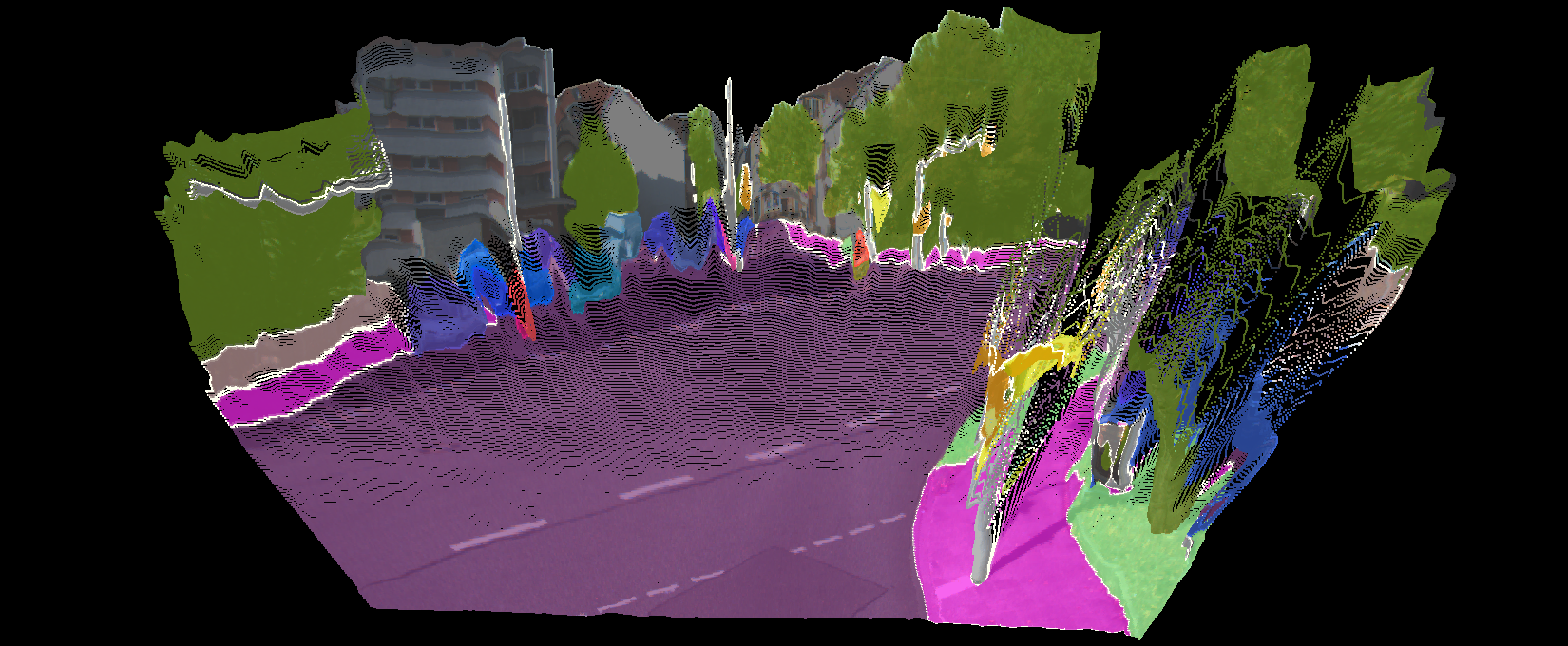}
\end{subfigure}
\par\medskip 
\begin{subfigure}[b]{0.35\textwidth}
  \centering
  \begin{tikzpicture}
    \node[anchor=south west,inner sep=0] at (0,0) {\includegraphics[width=\textwidth]{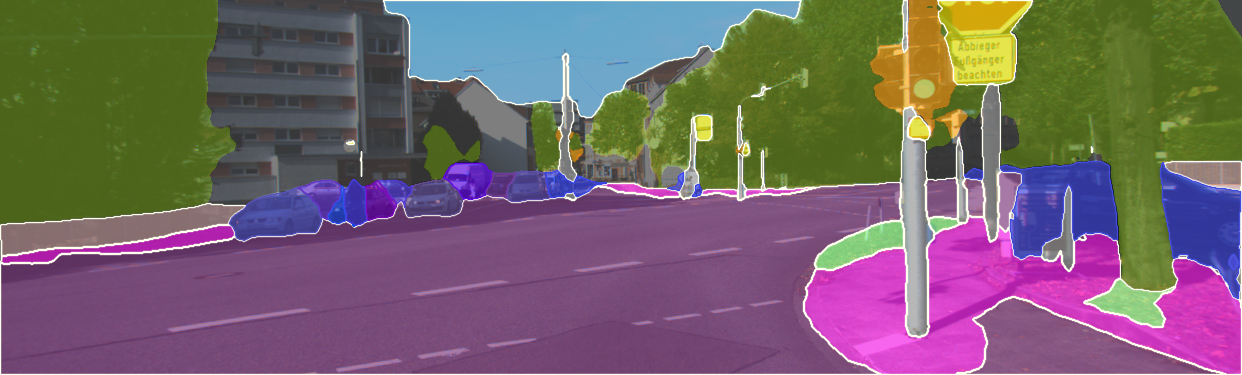}};
    \draw[red,thick] (1.15,0.45) rectangle (1.3,0.7);
    \draw[red,thick] (2.5,0.02) rectangle (4.26,0.35);
  \end{tikzpicture}
\end{subfigure}
\hfill
\begin{subfigure}[b]{0.35\textwidth}
  \centering
  \begin{tikzpicture}
    \node[anchor=south west,inner sep=0] at (0,0) {\includegraphics[width=\textwidth]{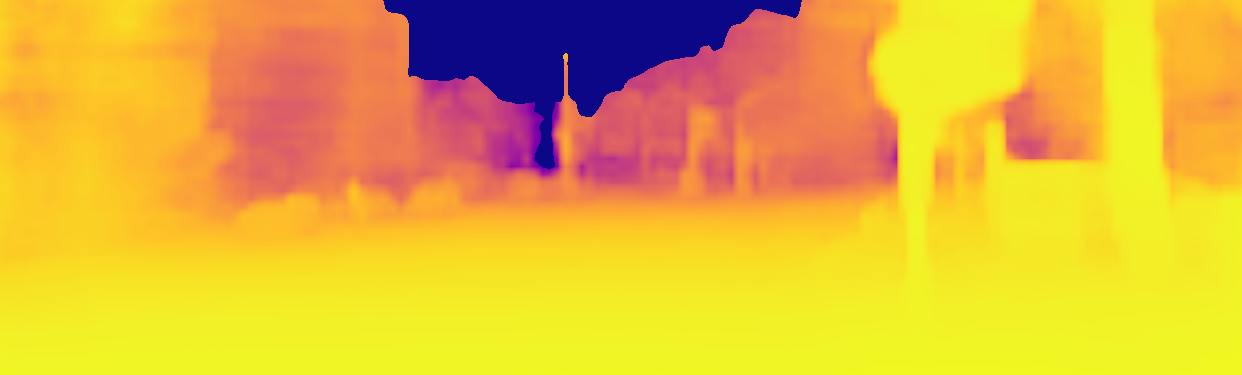}};
    \draw[red,thick] (2.3,0.55) rectangle (2.9,1.1);
  \end{tikzpicture}
\end{subfigure}
\hfill
\begin{subfigure}[b]{0.257\textwidth}
  \centering
  \includegraphics[width=\textwidth]{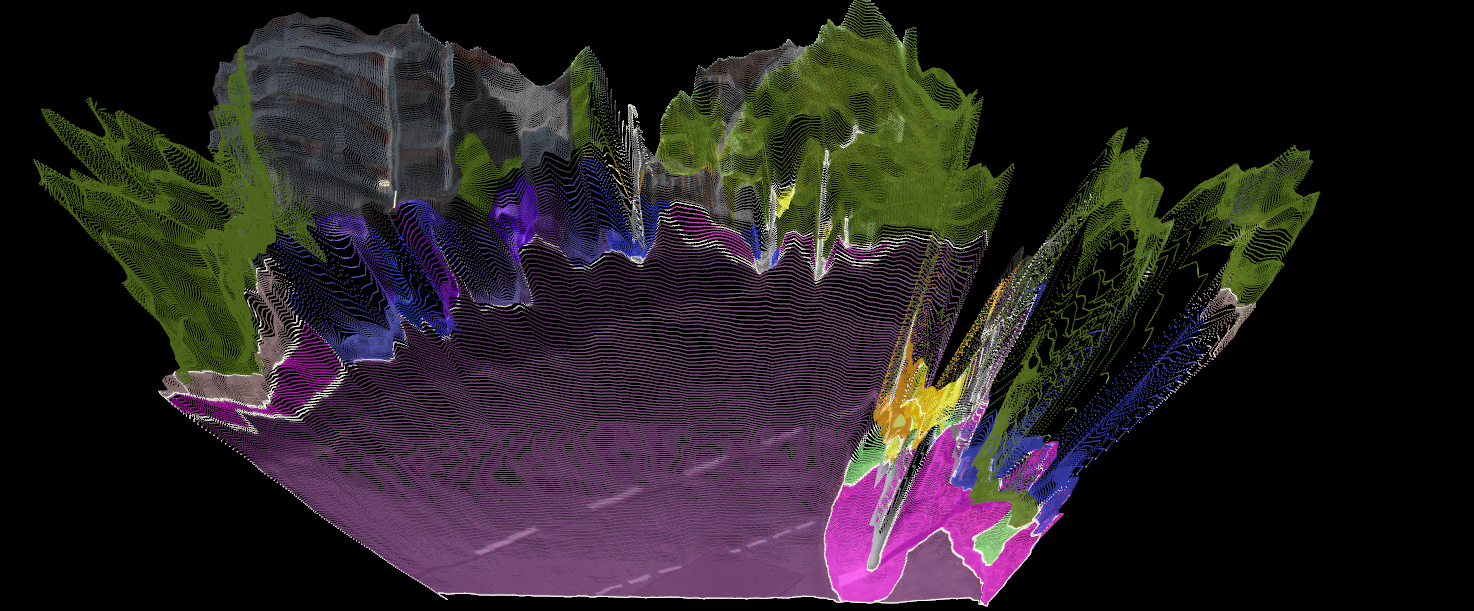}
\end{subfigure}
\caption{Comparison of our MGNiceNet (top) with MGNet~\cite{schoen2021mgnet} (bottom). MGNiceNet achieves more accurate segmentation of small objects, such as bicyclists, and large areas, such as sidewalks, as well as a more fine-grained depth prediction.}
\label{fig:mgnet_comparison}
\vspace{-5mm}
\end{figure}

\Cref{fig:qualitative} shows qualitative results of our method on images of the Cityscapes test dataset and the KITTI~\cite{geiger2013vision} Eigen test split~\cite{eigen2014depth}.
The first three rows per dataset show examples where our method performs well.
Our method produces high-quality panoptic segmentation and depth estimation predictions, showing its effectiveness.
The last row per dataset shows an example where the prediction is inaccurate.
For Cityscapes, our model wrongly predicts a bicycle on the back of the fire truck, leading to an incorrect depth prediction.
We argue that this corner case was not seen during training and is one opportunity for future improvement.
For KITTI, our model cannot accurately predict the segmentation mask and depth of the bridge across the street.
We argue that this is due to the usage of panoptic pseudo labels.
Since KITTI does not provide panoptic ground truth, our model is only trained using pseudo labels.
Thus, the panoptic prediction contains more incorrect segmentation masks than Cityscapes which also influences depth performance.
\Cref{fig:mgnet_comparison} shows a qualitative comparison of our MGNiceNet with our previous approach in monocular geometric scene understanding MGNet~\cite{schoen2021mgnet}.
Compared to MGNet, the unified approach improves both panoptic segmentation and depth estimation performance.
For panoptic segmentation, small objects, such as pedestrians or bicyclists, and large amorph regions, such as sidewalks or roads, are segmented more accurately.
For depth estimation, MGNiceNet especially shows improvements in capturing fine-grained details, such as poles.
We argue that the improvements are mainly due to our unified approach to monocular geometric scene understanding.

\section{Implementation Details}
\label{sec:supp_impl_details}
We implement our method in Pytorch~\cite{paszke2019pytorch} using the detectron2 framework~\cite{wu2019detectron}.
Since the ego-car region of Cityscapes does not adhere to the Lambertian assumptions in the photometric loss, we load pre-calculated ego-car masks and omit this region in the photometric loss calculation and during inference.
For panoptic segmentation post-processing, we follow RT-K-Net~\cite{schoen2023rt} settings.
In particular, we filter out masks with confidences below $\delta_s=0.3$ and an overlap below $\delta_o=0.6$.
Inference times are measured as an average of 500 forward passes through our model, including data-loading and post-processing.
Inference times are measured on a single NVIDIA Titan RTX GPU without TensorRT~\cite{TensorRT} optimization.
For comparison to state-of-the-art methods, we use inference times as stated in the respective publication or calculate inference times based on official code releases if available.

\section{Potential Negative Impact}
\label{sec:supp_negative_impact}
While our technical innovations do not appear to have ethical biases, trained models can reflect the inherent biases of the dataset used.
Hence, trained models should undergo an ethical review to ensure that predictions are not biased toward certain ethnic groups and that our method is not misused for applications such as illegal surveillance.
The two used datasets, Cityscapes and KITTI, both contain sensitive personal data, such as the faces of human subjects.
We used sensitive data carefully and according to the respective privacy protection agreement,~\eg, using anonymized images for our visualizations and demo videos if available.

\bibliographystyle{splncs04}
\bibliography{main}